\documentclass[11pt,a4paper]{article}
\usepackage[hyperref]{acl2018}
\usepackage{times}
\usepackage{latexsym}
\usepackage{multirow}
\usepackage{url}
\usepackage{float}

\aclfinalcopy 

\usepackage{graphicx}
\usepackage{enumitem}
\usepackage{color}
\usepackage{amsmath}
\usepackage{amssymb}
\usepackage{overpic}
\usepackage{wrapfig}
\usepackage{tabularx,ragged2e,booktabs,caption}
\usepackage{siunitx}
\usepackage{xcolor}
\usepackage{array}
\usepackage{arydshln}
\usepackage{natbib}

\usepackage{pifont}
\newcommand{\cmark}{\textcolor{red}{\ding{52}}}
\newcommand{\xmark}{\textcolor{blue}{\ding{56}}}

\newenvironment{shrinkeq}[1]
{ \bgroup
	\addtolength\abovedisplayshortskip{#1}
	\addtolength\abovedisplayskip{#1}
	\addtolength\belowdisplayshortskip{#1}
	\addtolength\belowdisplayskip{#1}}
{\egroup\ignorespacesafterend}

\graphicspath{{Result/}}

\title{Interactive Language Acquisition with One-shot Visual Concept Learning through a Conversational Game}

\author{Haichao Zhang$^{\dagger}$, Haonan Yu$^{\dagger}$, and  Wei Xu $^{\dagger}$$^{\S}$ \\
{\fontsize{11}{12} \selectfont \begin{tabular}{c}
	  $^{\dagger}$  {Baidu Research - Institue of Deep Learning, Sunnyvale USA}\\
	 $^{\S}$ {National Engineering Laboratory for Deep Learning Technology and Applications, Beijing China}\\
	\end{tabular}} \\
{\fontsize{11}{12}	\texttt{\{zhanghaichao,haonanyu,wei.xu\}@baidu.com}}\\
}

\date{}

\hypersetup{draft}

\begin{document}
	\maketitle
	\begin{abstract}
		Building intelligent agents that can communicate with and learn from humans in natural language is of great value.
		Supervised language learning is limited by the ability of capturing mainly the statistics of training data, and is hardly adaptive to new scenarios or flexible for acquiring new knowledge without inefficient retraining or catastrophic forgetting.
		We highlight the perspective that conversational interaction serves as a natural interface  both for  language learning and  for novel knowledge acquisition and  propose a joint imitation and reinforcement approach for grounded language learning through an interactive conversational game. The agent trained with this approach is able to actively acquire information by asking questions about novel objects and use the just-learned knowledge in subsequent conversations in a one-shot fashion. 
		Results compared with other methods verified the effectiveness of the proposed approach.
	\end{abstract}
	
	\section{Introduction}\label{sec:Intro}
	Language is  one of the most natural forms of communication for human and is typically viewed as fundamental to human intelligence; therefore it is crucial for an intelligent  agent to be able to use language to communicate with human  as well.
	While supervised training with deep neural networks has led to encouraging progress in language learning,  it suffers from the problem of capturing mainly the statistics of training data, and from a lack of adaptiveness to new scenarios and being flexible for acquiring new knowledge without inefficient retraining or  catastrophic forgetting.
	Moreover, supervised training of deep neural network models needs a large number of training samples， while many interesting applications require rapid learning from a small amount of data, which poses an even greater challenge to the supervised setting.
	
	In contrast, humans learn in a way  very different from the supervised setting~\cite{BFSkinner,nature_lang}.
	First, humans act upon the world and learn from the consequences of their actions~\cite{BFSkinner,nature_lang,rl_lang_learning}.
	While for mechanical actions such as movement, the consequences mainly follow geometrical and mechanical principles, for language, humans act by speaking, and the consequence is typically a response in the form of verbal and other behavioral feedback (\emph{e.g.}, nodding) from the conversation partner (\emph{i.e.}, teacher). 
	These types of feedback typically contain  informative signals on how to improve language skills in  subsequent conversations and play an important role in humans' language acquisition process~\cite{nature_lang,rl_lang_learning}.
	Second, humans have shown a celebrated ability to learn  new concepts from small amount of data~\cite{oneshot_word}.
	From even just one example, children seem to be able to make inferences
	and draw plausible boundaries
	between concepts, demonstrating the ability of one-shot learning~\cite{lake_oneshot}.

	The language acquisition process and the one-shot learning ability of human beings are both impressive as a manifestation of human intelligence, and are inspiring for designing novel settings and algorithms for computational language learning. 
	In this paper, we leverage conversation  as both  an interactive environment for language learning~\cite{BFSkinner} and a natural interface for acquiring new knowledge~\cite{conversational_learning}.
	We propose an approach for interactive language acquisition with  one-shot concept learning ability.
	The proposed approach allows an agent to learn grounded language from scratch,  acquire the transferable skill of actively seeking and memorizing information about novel objects, and develop the one-shot learning ability, purely through conversational interaction with a teacher.

	\section{Related Work}
	\label{sec:related_works}
	{\flushleft \textbf{Supervised Language Learning}.} Deep neural network-based language learning has seen great success on many applications, including  machine translation~\cite{cho-al-emnlp14}, dialogue generation~\cite{lang_generation,HRNN_dialogue16}, image captioning and visual question answering~\cite{mao2014deep, VQA}.
	For training, a large amount of labeled data is needed, requiring significant  efforts to collect.
	Moreover, this setting essentially captures the statistics of training data and does not respect the interactive nature of language learning,
	rendering it less flexible for acquiring new knowledge without retraining or  forgetting~\cite{book_dialogue_supervised}.

	{\flushleft \textbf{Reinforcement Learning for Sequences}.}
	Some recent studies used  reinforcement learning (RL) to tune the performance of a pre-trained language model  according to certain metrics~\cite{RL_Seq_ICLR15, AC_Seq,RL_Dialogue,SeqGAN}.
	Our work is also related to RL in  natural language action space~\cite{Lan_Action_Space} and  shares a similar motivation with~\citet{DBLL} and \citet{dialogue_question}, 
	which explored language learning through pure textual dialogues.
	However, in these works~\cite{Lan_Action_Space, DBLL, dialogue_question},
	a set of candidate sequences is provided and the action  is to select one from the  set.
	Our main focus is rather on learning language from scratch: the agent has to learn to \emph{generate} a sequence action rather than to simply \emph{select}
	one from a provided candidate set.
	
	\vspace{-0.05in}
	{\flushleft \textbf{Communication and Emergence of Language}.} 
	Recent studies have examined learning to communicate~\cite{RL_Com, BP_Com} and invent language~\cite{Multi_Agent_Lan,Emergence_Lan}. The emerged language needs to be interpreted by humans via post-processing~\cite{Emergence_Lan}.
	We, however, aim to achieve language learning from the dual  perspectives of understanding and generation, and the speaking action of the agent is readily understandable without any post-processing.
	Some studies on language learning have used a guesser-responder setting in which the guesser tries to achieve the final goal (\emph{e.g.}, classification) by collecting additional information through asking the responder questions~\cite{Dialogue_Deepmind, visdial_rl}. 
	These works try to optimize the question being asked to help the guesser achieve the final goal, while we focus on transferable speaking and one-shot ability.

	{\flushleft \textbf{One-shot Learning and Active Learning}.}
	One-shot learning has been investigated in some recent works~\cite{lake_oneshot,MANN,active_oneshot}. 
	The memory-augmented network~\cite{MANN}  stores visual representations mixed with  ground truth class labels in an external memory for one-shot learning. A class label is always provided following the presentation of an image; thus the agent receives information from the teacher in a passive way. 
	\citet{active_oneshot} present efforts toward active learning, using a vanilla recurrent neural network (RNN) without an external memory. Both lines of study focus on image classification only, meaning the class label is directly provided for memorization. In contrast, we target language and one-shot learning via conversational interaction, and the learner has to learn to extract important information from the  teacher's sentences for memorization.

	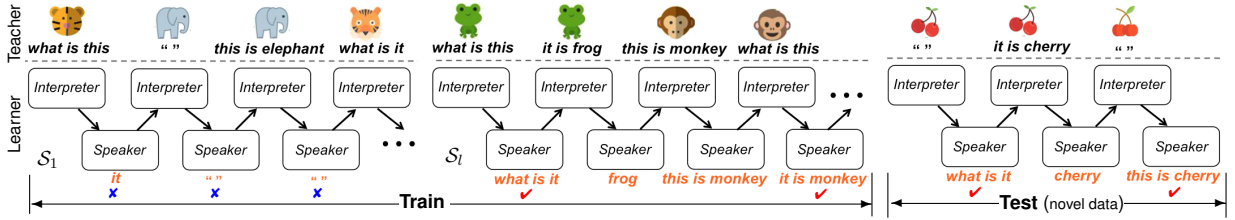
\begin{figure*}[t]
		\hspace{0.05in}
		\begin{overpic}[viewport = 3 5 700 355, clip, height = 2.6cm]{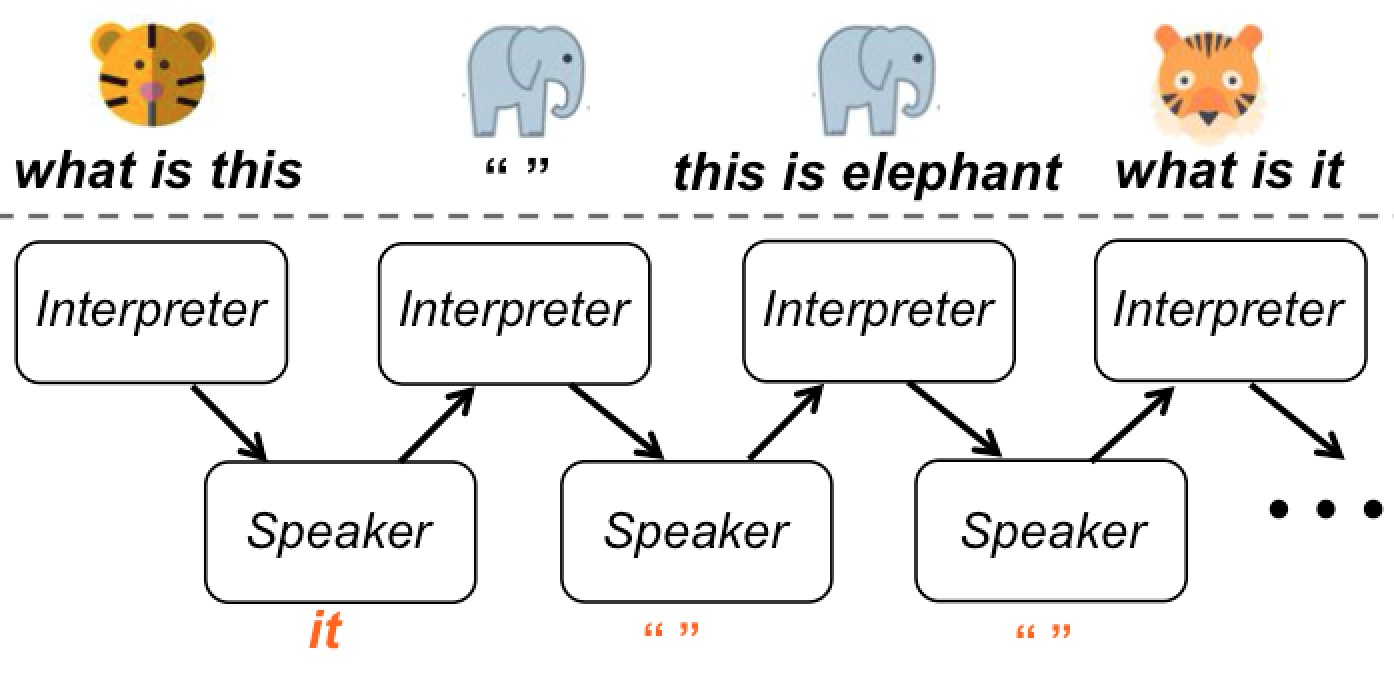}
			\put(3,7){{\colorbox{white}{\textcolor{black}{\scalebox{.7}{$\mathcal{S}_{1}$}}}}}
			\put(1,-5.5){\line(0,1){10}}
			\put(100,-3){\vector(1,0){115}}
			\put(100,-3){\vector(-1,0){98.5}}
			\put(95,-4){\sffamily \small {\colorbox{white}{\textcolor{black}{{\scalebox{.8}{{\textbf{Train} }}}}}}}
			\put(-4.5,33){\sffamily \small {\textcolor{black}{{\scalebox{.7}{\rotatebox{90}{Teacher}}}}}}		\put(-4.5,10){\sffamily \small {\textcolor{black}{{\scalebox{.7}{\rotatebox{90}{Learner}}}}}}
			\put(21,-1.5){\textcolor{blue}{\scalebox{.5}{{\xmark}}}}				\put(47,-1.5){\textcolor{blue}{\scalebox{.5}{{\xmark}}}}					\put(73,-1.5){\textcolor{blue}{\scalebox{.5}{{\xmark}}}}
		\end{overpic}
		\hspace{0.005in}
		\begin{overpic}[viewport = 1 5 780 355, clip,height = 2.6cm]{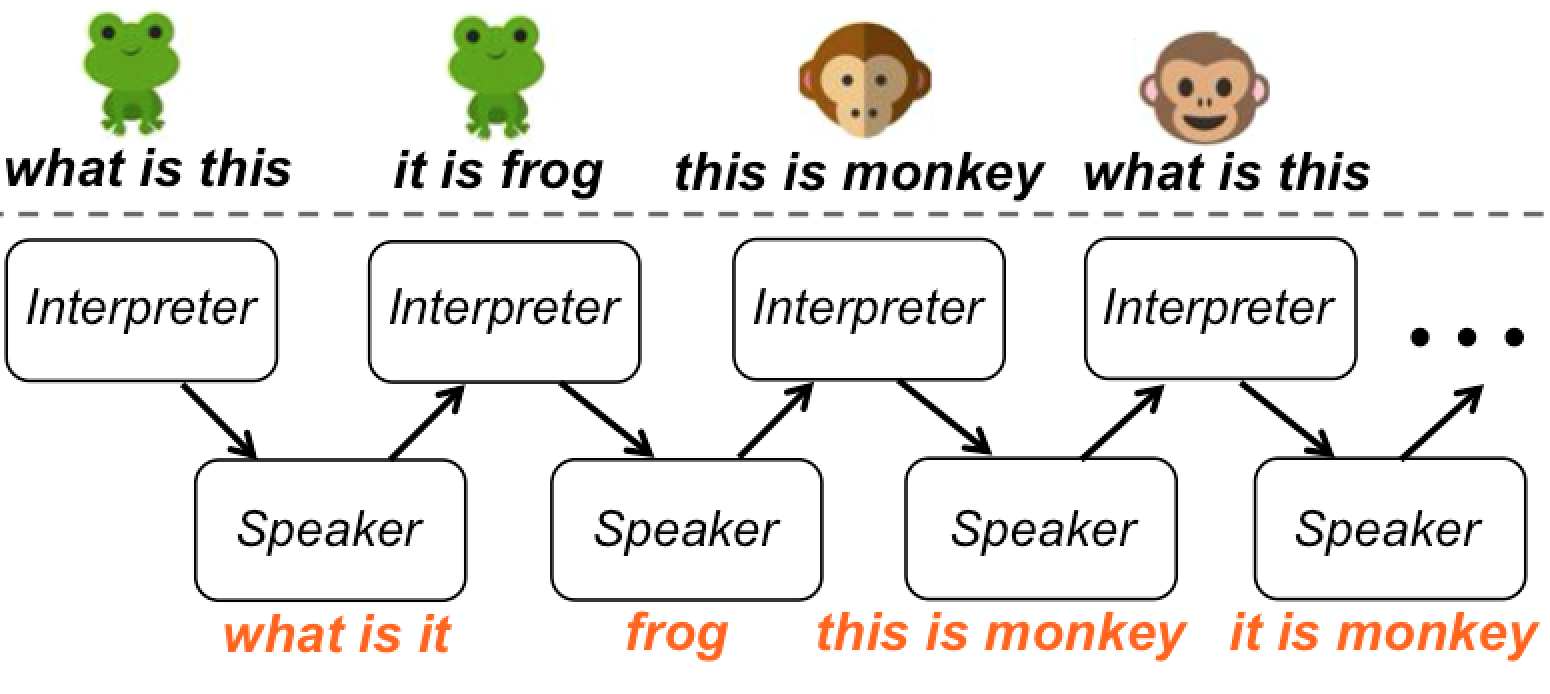}
			\put(3,7){{\colorbox{white}{\textcolor{black}{\scalebox{.7}{$\mathcal{S}_{l}$}}}}}
			\put(100,-5.5){\line(0,1){10}}
			\put(20,-1.5){\textcolor{red}{\scalebox{.5}{{\cmark}}}}				
			\put(87,-1.5){\textcolor{red}{\scalebox{.5}{{\cmark}}}}
		\end{overpic}
		\hspace{0.02in}
		\begin{overpic}[viewport = 1 1 600 351, clip,height = 2.6cm]{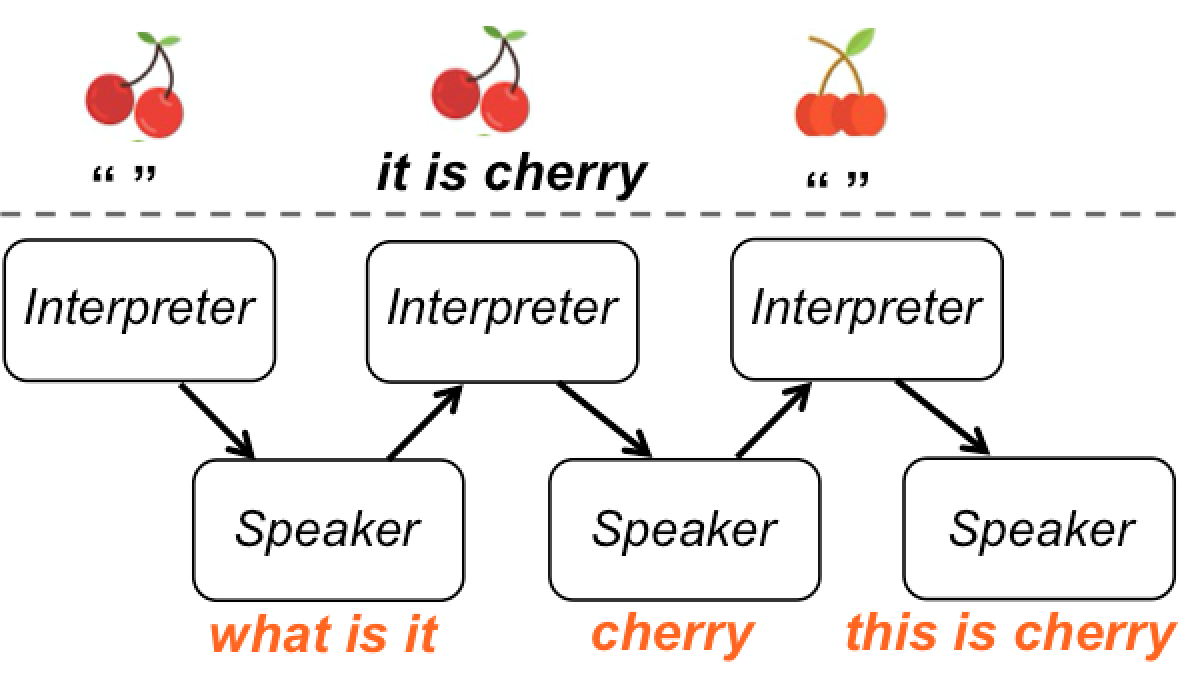}
			\put(0,-7.55){\line(0,1){13.4}}
			\put(100,-7.55){\line(0,1){13.4}}
			\put(35,-3.3){\vector(-1,0){34}}
			\put(50,-3.3){\vector(1,0){50}}
			\put(33,-5){\sffamily \small {\colorbox{white}{\textcolor{black}{{\scalebox{.8}{{\textbf{Test} (\textsf{\scriptsize{novel data}})}}}}}}}
			\put(24,-1.5){\textcolor{red}{\scalebox{.5}{{\cmark}}}}				\put(83,-1.5){\textcolor{red}{\scalebox{.5}{{\cmark}}}}
		\end{overpic}
		\caption{\textbf{Interactive language and one-shot concept learning.} Within a session $\mathcal{S}_l$, the teacher may ask questions, answer learner's questions, make statements, or say nothing. The teacher
			also provides reward feedback based on learner's responses as (dis-)encouragement.
			The learner alternates between interpreting teacher's sentences and generating a response through \emph{interpreter} and \emph{speaker}.
			\textbf{Left:} Initially, the learner can barely say anything meaningful. \textbf{Middle:} Later it can produce meaningful responses for interaction.  
			\textbf{Right:}
			After training, when confronted with an image of \emph{cherry}, which is a novel class that the learner never saw before during training, the learner can ask a question about it (``\emph{what is it}'') and generate a correct statement (``\emph{this is cherry}'') for another instance of cherry after only being taught once. 
		}
		\label{fig:setting}
		\vspace{-0.1in}
	\end{figure*}

	\section{The Conversational Game}
	\label{sec:game}
	We construct a conversational game inspired by experiments on language  development in infants  from cognitive science~\cite{game}.
	The game is implemented with the \textsc{xworld} simulator~\cite{Nav, L2T} and is publicly available online.\footnote{\fontsize{8}{12}\selectfont\url{https://github.com/PaddlePaddle/XWorld}}
	It provides an environment for the agent\footnote{We use the term \emph{agent} interchangeably with \emph{learner}.} to learn language and develop the one-shot learning ability.
	One-shot learning here means that during test sessions, no further training happens to the agent and it has to answer teacher's questions correctly about novel images of never-before-seen classes after being taught only once by the teacher, as illustrated in Figure~\ref{fig:setting}.
	To succeed in this game, the agent has to learn to 1) speak by generating sentences, 
	2) extract and memorize useful information with only one exposure and use it in subsequent conversations,
	and 3) behave adaptively according to context and its own knowledge (\emph{e.g.}, asking questions about unknown objects and answering questions about something known), all achieved through interacting with the teacher.
	This makes our game distinct from other seemingly relevant games, in which the agent cannot speak~\cite{wang2016games} or ``speaks'' by \emph{selecting} a candidate from a provided set~\cite{Lan_Action_Space,DBLL,dialogue_question}  rather than \emph{generating} sentences by itself, or games mainly focus on slow learning~\cite{visdial_rl,Dialogue_Deepmind}  and falls short on one-shot learning.

	In this game, sessions ($\mathcal{S}_l$) are randomly instantiated during interaction.
	Testing sessions are constructed with a separate dataset with concepts that never appear before during training to  evaluate the language and one-shot learning ability.  
	Within a session, the teacher randomly selects an object and interacts with the learner about the object by  randomly 1)~posing a question (\emph{e.g.}, ``\emph{what is this}"), 2)~saying nothing (\emph{i.e.}, \emph{``''}) or 3)~making a statement (\emph{e.g.}, ``\emph{this is monkey}"). 
	When the teacher asks a question or says nothing, i)  if the learner raises a question, the teacher will provide a statement about the object asked (\emph{e.g.}, ``\emph{it is frog}")  with a \emph{question-asking reward} ($+0.1$); ii) if the learner says nothing, the teacher will still provide an answer (\emph{e.g.}, ``\emph{this is elephant}") but with an \emph{incorrect-reply reward} ($-1$) to discourage the learner from remaining silent; iii) for all other incorrect responses from the learner, the teacher will provide an \emph{incorrect-reply reward} and move on to the next random object for interaction. 
	When the teacher generates a statement,  the learner will receive
	no reward if a correct statement is generated otherwise an \emph{incorrect-reply reward} will be given. 
	The session ends if the learner answers the teacher's question correctly, generates a correct statement when the teacher says nothing (receiving a \emph{correct-answer reward}~$+1$), or when the maximum number of steps is reached. 
	The sentence from teacher at each time step is generated using a context-free grammar as shown in Table~\ref{tab:grammar}.

	A success is reached if the learner behaves correctly  during the whole session: asking questions about novel objects, generating answers when asked, and making statements when the teacher says nothing about objects that have been taught within the session. Otherwise it is a \mbox{failure}.
	
	\vspace{-0.05in}
	\begin{table}[h]
		\footnotesize
		\caption{{Grammar for the teacher's sentences.}} 
		\vspace{-0.1in}
		\label{tab:grammar}
		\begin{tabular}{p{1.3cm}@{$\rightarrow$ \,}p{5.5cm}}
			\hline	
		 start & question  $\vert$ silence $\vert$ statement\\
		 question & Q1 $\vert$ Q2 $\vert$ Q3 \\
			 silence & `` '' \\
		 statement & A1 $\vert$ A2 $\vert$ A3 $\vert$ A4 $\vert$ A5 $\vert$ A6 $\vert$ A7 $\vert$ A8 \\
		 Q1 & ``\emph{what}'' \\
		 Q2 & ``\emph{what}'' M \\
		 Q3 & ``\emph{tell what}'' N \\
		 M & ``\emph{is it}'' $\vert$ ``\emph{is this}'' $\vert$ ``\emph{is there}'' $\vert$
						  ``\emph{do you see}'' $\vert$ ``\emph{can you see}'' $\vert$
						  ``\emph{do you observe}'' $\vert$ ``\emph{can you observe}'' \\
		 N & ``\emph{it is}'' $\vert$ ``\emph{this is}'' $\vert$ ``\emph{there is}'' $\vert$ ``\emph{you see}'' $\vert$ ``\emph{you can see}'' $\vert$ ``\emph{you observe}'' $\vert$ ``\emph{you can observe}''\\
		 A1 & G \\
		 A2 & ``\emph{it is}'' G \\
		 A3 & ``\emph{this is}'' G \\
		 A4 & ``\emph{there is}'' G \\
		 A5 & ``\emph{i see}'' G \\
		 A6 & ``\emph{i observe}'' G \\
		 A7 & ``\emph{i can see}'' G \\
		 A8 & ``\emph{i can observe}'' G \\
		 G  &  object name \\
			\hline
		\end{tabular}
	\end{table}

	\section{Interactive Language Acquisition via Joint Imitation and Reinforcement}
	\label{sec:model}
	{\flushleft\textbf{Motivation}}.
	The goal is to 
	\emph{learn to converse and develop the one-shot learning ability by conversing with a teacher and improving from teacher's feedback}. 
	We propose to use a \emph{joint imitation and reinforce} approach to achieve this  goal.
	\mbox{\emph{{Imitation}}} helps the agent to develop the basic ability to generate sensible sentences. 
	As learning is done by observing the teacher's behaviors during conversion,
	the agent essentially imitates the teacher from a \emph{third-person perspective}~\citep{third_person} rather than imitating an expert agent who is conversing with the teacher~\cite{visdial_rl,Dialogue_Deepmind}.
	During conversations, the agent perceives sentences and images  without any explicit labeling of ground truth answers, and it has to learn to make sense of raw perceptions, extract useful information, and save it for later use when generating an answer to teacher's question. 	
	While it is tempting to purely imitate the teacher, the agent trained this way only develops \emph{echoic behavior}~\cite{BFSkinner}, \emph{i.e.},  mimicry.
	\emph{{Reinforce}} leverages  confirmative feedback from the teacher for learning to converse adaptively beyond mimicry by adjusting the action policy. It enables the learner to use the acquired speaking ability and adapt it according to reward feedback.  This is analogous to some views on the babies' language-learning process that babies use the acquired speaking skills by trial and error with parents and improve according to the consequences of speaking actions~\cite{BFSkinner, rl_lang_learning}. 
	The fact that babies don't fully develop the speaking capabilities without the ability to hear~\cite{hearing}, and that it is hard to make a meaningful conversation with a trained parrot signifies the importance of both imitation and reinforcement in language learning.
		
	{\flushleft\textbf{Formulation}}. The agent's response can be modeled as a sample from a probability distribution over the possible sequences. Specifically, for one session, given the visual input $\mathbf{v}^t$ and  conversation history $\mathcal{H}^t\!\!\!=\!\!\!\{\mathbf{w}^1, \mathbf{a}^1, \cdots, \mathbf{w}^t\}$, the agent's response $\mathbf{a}^t$ can be generated by sampling from a distribution  of the speaking action $\mathbf{a}^t \!\!\!\sim\!\!\! p_{\theta}^{\rm S}(\mathbf{a}|\mathcal{H}^t, \mathbf{v}^t)$.
	The agent interacts with the teacher by outputting the utterance $\mathbf{a}^t$  and receives feedback from the teacher in the next step, with 
	$\mathbf{w}^{t+1}$ a sentence as verbal feedback 
	and $r^{t+1}$  reward feedback (with positive values as encouragement while negative values as discouragement, according to $\mathbf{a}^t$, as described in Section~\ref{sec:game}). 
	Central to the goal is learning $p_{\theta}^{\rm S}(\cdot)$.
	We formulate the problem as the minimization of a cost function as:
	\begin{shrinkeq}{-0.6ex}
		\begin{eqnarray}
		\label{cost_function}
		\nonumber
		\textstyle{
			\mathcal{L}_{\theta}\!\! = \!\!\underbrace{\mathbb{E}_{\mathcal{W}}{\textstyle\big[\!-\!\! \sum_t\log p_{\theta}^{\rm I}(\mathbf{w}^{t}|\cdot)\big]}}_{\rm Imitation \; \mathcal{L}^{\rm I}_{\theta}} \!\!+\!\!\underbrace{\mathbb{E}_{p^{\rm S}_{\theta}} {\big[\!-\!\! {\textstyle\sum_t} [\gamma]^{t-1} \cdot r^{t}\big]}}_{\rm Reinforce \; \mathcal{L}^{\rm R}_{\theta}}}
		\end{eqnarray}
	\end{shrinkeq}
	where $\mathbb{E}_{\mathcal{W}}(\cdot)$ is the expectation over all the sentences $\mathcal{W}$  from teacher,
	$\gamma$ is a reward discount factor, and  
	$[\gamma]^{t}$ denotes the exponentiation over $\gamma$.
	While the imitation term learns directly the predictive distribution $p_{\theta}^{\rm I}(\mathbf{w}^{t}|\mathcal{H}^{t-1}, \mathbf{a}^t)$, it contributes to $p_{\theta}^{\rm S}(\cdot)$  through \emph{parameter sharing} between them.

	\begin{figure*}[t]
		\centering
		\begin{overpic}[viewport = 35 340 540 575, clip, height = 5.5cm]{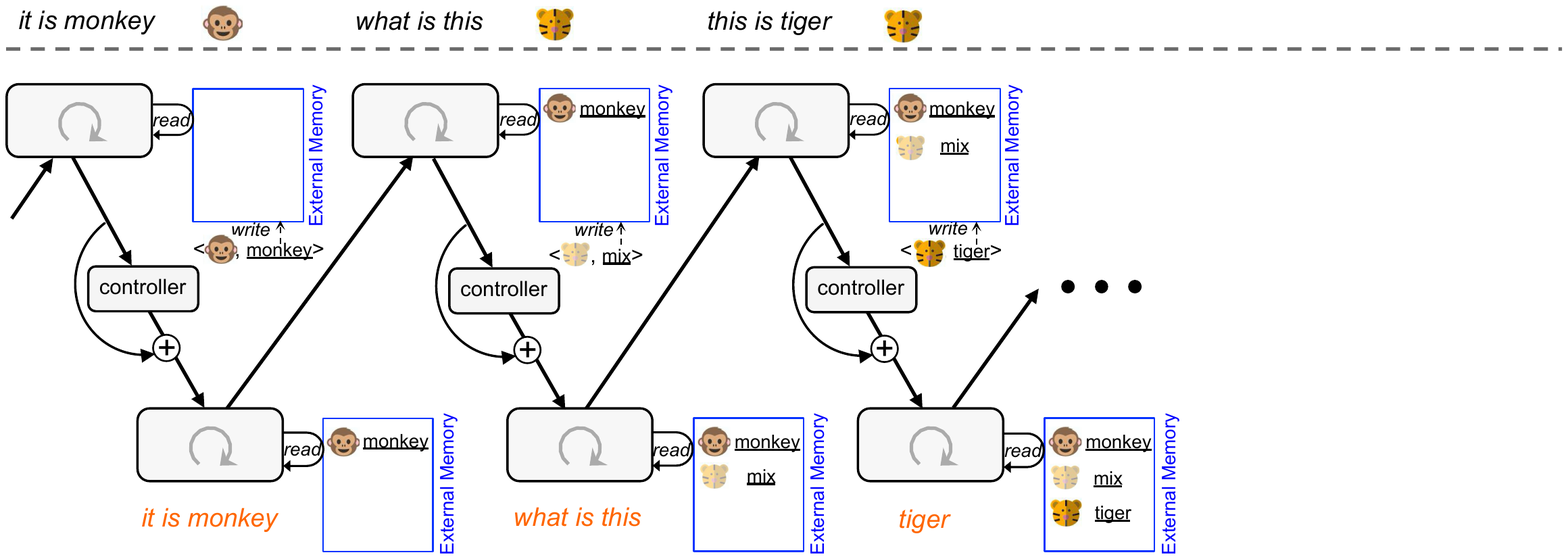}
			\put(0,42.){{\colorbox{white}{\textcolor{black}{\scalebox{0.7}{$t \longmapsto$}}}}}
			\put(29.5,42.){{\colorbox{white}{\textcolor{black}{\scalebox{0.7}{$t+1 \longmapsto$}}}}}
			\put(59.9,42.){{\colorbox{white}{\textcolor{black}{\scalebox{0.7}{$t+2 \longmapsto$}}}}}
			
			\put(13.5,45){{{\textcolor{black}{\scalebox{.7}{$\mathbf{w}^{t}$}}}}}
			\put(21,45){{{\textcolor{black}{\scalebox{.7}{$\mathbf{v}^{t}$}}}}}
			\put(0,29){{\colorbox{white}{\textcolor{black}{\scalebox{.7}{$\mathbf{h}_{\rm last}^{t-1}$}}}}}
			\put(28,29){{\colorbox{white}{\textcolor{black}{\scalebox{.7}{$\mathbf{h}_{\rm last}^{t}$}}}}}
			\put(9,30.5){{\colorbox{white}{\textcolor{black}{\scalebox{.7}{$\mathbf{h}_{\rm I}^{t}$}}}}}		\put(13.5,32.8){{\colorbox{white}{\textcolor{black}{\scalebox{.7}{$\mathbf{r}_{\rm I}^{t}$}}}}}
			\put(12,13.5){{\colorbox{white}{\textcolor{black}{\scalebox{.7}{$\mathbf{c}^{t}$}}}}}
			\put(24.5,5){{\colorbox{white}{\textcolor{black}{\scalebox{.7}{$\mathbf{r}_s^{t}$}}}}}
			\put(17,-0.5){{\colorbox{white}{\textcolor{black}{\scalebox{.7}{$\mathbf{a}^{t}$}}}}}
			
			\put(-2,30){\sffamily \small {\textcolor{black}{{\scalebox{.8}{\rotatebox{90}{Interpreter}}}}}}
			\put(-2,5){\sffamily \small {\textcolor{black}{{\scalebox{.8}{\rotatebox{90}{Speaker}}}}}}
			\put(-1,25){\vector(0,1){5}}
			\put(-1,26.5){\vector(0,-1){13}}
			\put(-1,15.65){\sffamily \scriptsize {\textcolor{black}{{\scalebox{.8}{\rotatebox{90}{share parameter}}}}}}
			
			\put(-3.5,41){\sffamily \small{\textcolor{black}{{\scalebox{.7}{\rotatebox{90}{Teacher}}}}}}		\put(-3.5,20){\sffamily \small {\textcolor{black}{{\scalebox{.7}{\rotatebox{90}{Learner}}}}}}
			
			\put(-0, -0.5){\sffamily \scriptsize {\textcolor{black}{{(a)}}}}
		\end{overpic}
		\hspace{0.3in}
		\begin{overpic}[viewport = 5 5 343 660, clip, height = 5.5cm]{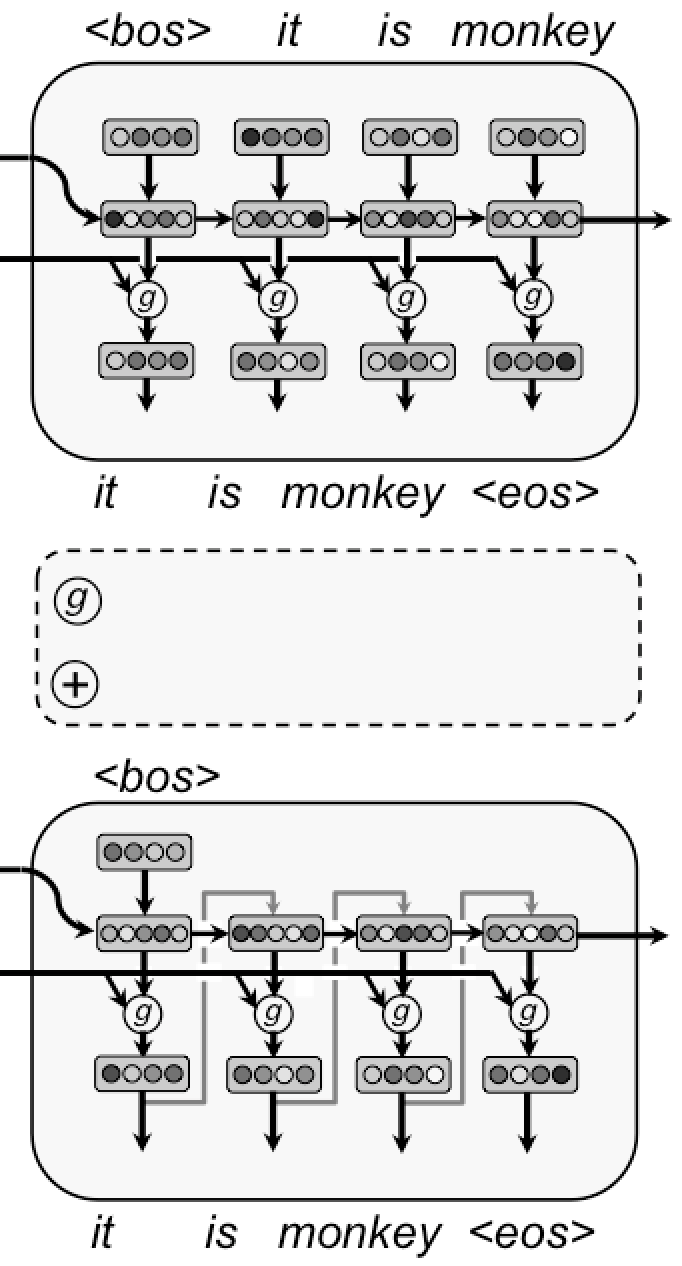}
			\put(-14.,62){\sffamily \small {\textcolor{black}{{\scalebox{.7}{\rotatebox{90}{Interpreter-RNN}}}}}}
			\put(-14,5){\sffamily \small {\textcolor{black}{{\scalebox{.7}{\rotatebox{90}{Speaker-RNN}}}}}}
			\put(-13,40){\vector(0,1){22}}
			\put(-13,42){\vector(0,-1){12}}
			\put(-13,38){\sffamily \scriptsize {\textcolor{black}{{\scalebox{.8}{\rotatebox{90}{share para.}}}}}}
			\put(8,49.5){{\sffamily \scalebox{.5} {\textcolor{black}{{memory-RNN fusion gate}}}}}
			\put(11,43.5){{\sffamily \scalebox{.5} {\textcolor{black}{{additive aggregation}}}}}
			
			\put(-9,82){{\colorbox{white}{\textcolor{black}{\scalebox{.7}{$\mathbf{h}_{\rm last}^{t-1}$}}}}}		\put(-5,73){{\colorbox{white}{\textcolor{black}{\scalebox{.7}{$\mathbf{r}_{\rm I}^{t}$}}}}}
			\put(50,81.5){{\colorbox{white}{\textcolor{black}{\scalebox{.7}{$\mathbf{h}_{\rm I}^{t}$}}}}}
			
			\put(-5,29){{\colorbox{white}{\textcolor{black}{\scalebox{.7}{$\mathbf{c}^{t}$}}}}}		\put(-5,20){{\colorbox{white}{\textcolor{black}{\scalebox{.7}{$\mathbf{r}_{s}^{t}$}}}}}
			\put(50,27){{\colorbox{white}{\textcolor{black}{\scalebox{.7}{$\mathbf{h}_{\rm last}^{t}$}}}}}

			\put(-7,-1){\sffamily \scriptsize {\textcolor{black}{{(b)}}}}
		\end{overpic}
		\caption{\textbf{Network structure.} (a) Illustration of the overall architecture. At each time step, the learner uses the interpreter module to encode the teacher's sentence. The visual perception is also encoded and used as a key to retrieve information from the external memory. The last state of the interpreter-RNN will be passed through a controller. The  controller's output will be added to the input and used  as the initial state of the speaker-RNN.
			The interpreter-RNN will update the external memory with an importance (illustrated with transparency) weighted information extracted from the perception input.
		`Mix' denotes a mixture of word embedding vectors.	
			(b) The structures of the interpreter-RNN (top) and the speaker-RNN (bottom). The interpreter-RNN and speaker-RNN share parameters.}
		\label{fig:network_structure}
		\vspace{-0.1in}
	\end{figure*}

	\vspace{-0.05in}
	{\flushleft \textbf{Architecture}.} 
	The learner comprises four major components:  \emph{external memory}, \emph{interpreter}, \emph{speaker}, and \emph{controller}, as shown in Figure~\ref{fig:network_structure}. 
	\emph{External memory} is flexible for storing and retrieving information~\cite{NTM, MANN}, making it a natural  component of our network for one-shot learning.
	The \emph{interpreter} is responsible for interpreting the teacher's sentences, extracting information from the perceived signals, and saving it to  the external memory.
	The \emph{speaker} is in charge of generating sentence responses with reading access to the external memory. The response could be a question asking for information or a statement answering a teacher's question, leveraging the information stored in the external memory.  The \emph{controller} modulates the behavior of the speaker to generate responses according to context (\emph{e.g.}, the learner's knowledge status).

	At time step $t$, the \emph{interpreter} uses an interpreter-RNN to encode the input sentence $\mathbf{w}^t$ from the teacher as well as historical conversational information into a state vector $\mathbf{h}_{\rm I}^t$.  
	$\mathbf{h}_{\rm I}^t$ is then passed through a residue-structured network, which is an identity mapping augmented with a learnable controller $f(\cdot)$ implemented with fully connected layers for producing $\mathbf{c}^t$.  Finally, $\mathbf{c}^t$ is used as the initial state of the speaker-RNN for generating the response $\mathbf{a}^t$.
	The final state $\mathbf{h}^t_{\rm last}$ of the speaker-RNN  will be used as the initial state of the interpreter-RNN at the next time step.
	
	\vspace{-0.05in}
	\subsection{Imitation with Memory Augmented Neural Network for Echoic Behavior}
	\label{sec:imitation}
	
	The teacher's way of speaking provides a source for the agent to imitate. For example, the syntax for composing a  sentence is a useful skill the agent can learn from the teacher's sentences, which could benefit both \emph{interpreter} and \emph{speaker}.
	Imitation is achieved by predicting teacher's future sentences with \emph{interpreter} and  parameter sharing between \emph{interpreter} and \emph{speaker}. 
	For prediction, we can represent the  probability of the next sentence $\mathbf{w}^{t}$ conditioned on the image $\mathbf{v}^{t}$ as well as previous sentences from both the teacher and the learner $\{\mathbf{w}^{1}, \mathbf{a}^{1}, \cdots, \mathbf{w}^{t-1}, \mathbf{a}^{t-1}\}$ as
	\begin{shrinkeq}{-0.3ex}
	\begin{eqnarray}
	\label{I_prob_new}
	\begin{split}
	&p_{\theta}^{\rm I}(\mathbf{w}^{t}| \mathcal{H}^{t-1}, \mathbf{a}^{t-1}, \mathbf{v}^{t}) \\
	&={\textstyle \prod_{i}}  \; p_{\theta}^{\rm I}(w^{t}_i|w_{1:i-1}^{t}, \mathbf{h}^{t-1}_{\rm last}, \mathbf{v}^{t}),
	\end{split}
	\end{eqnarray} 
	\end{shrinkeq}
	where $\mathbf{h}^{t-1}_{\rm last}$ is the last state of the RNN at time step $t\!{\displaystyle-}\!1$ as the summarization of $\{\mathcal{H}^{t-1}, \mathbf{a}^{t-1}\}$ (\emph{c.f.}, Figure~\ref{fig:network_structure}), and
	$i$ indexes words within a sentence.
	
	It is natural to model the probability of the \mbox{$i$-th} word in the $t$-th sentence with an RNN, where the sentences up to $t$ and words up to $i$ within the $t$-th sentence are captured by a fixed-length  state vector $\mathbf{h}_i^{t} \!\!= \!\!{\rm RNN} (\mathbf{h}_{i-1}^{t}, {w}_i^{t})$. 
	To incorporate knowledge learned and stored in the external memory, 
	the generation of the next word is \emph{adaptively} based on i) the predictive distribution of the next word from the state of the  RNN to capture the \emph{syntactic structure of sentences}, and ii) the information from the external memory to represent the previously learned knowledge, via a fusion gate $g$:
	\begin{equation}
	\label{eq:fuse}
	\centering
	p_{\theta}^{\rm I}(w^{t}_i|\mathbf{h}_i^{t}, \mathbf{v}^{t})  = (1-g) \cdot p_{\mathbf{h}} + g \cdot p_{\mathbf{r}},
	\end{equation}
	where $p_{\mathbf{h}}\!\! =\!\! {\rm softmax} \big(\mathbf{E}^{\mathsf{T}} f_{\rm MLP}(\mathbf{h}^{t}_i)\big)$ and $p_{\mathbf{r}} \!\!=\!\! {\rm softmax} \big(\mathbf{E}^{\mathsf{T}}\mathbf{r}\big)$. 
	$\mathbf{E} \!\!\in\!\! \mathbb{R}^{d \times k}$ is the  word embedding table, with  $d$ the embedding dimension   and $k$ the  vocabulary size.
	$\mathbf{r}$ is a vector read out from  the external memory using a visual key as detailed in the next section.
	$f_{\rm MLP}(\cdot)$ is a multi-layer Multi-Layer Perceptron (MLP) for bridging the semantic gap between the RNN state space and the word embedding space. 
	The fusion gate $g$ is computed as
	$g\!\! =\!\! f(\mathbf{h}^{t}_i, c)$,
	where $c$ is the confidence score  $c\!\!=\!\!\max (\mathbf{E}^{\mathsf{T}} \mathbf{r})$, and a well-learned concept should have a large score by design (Appendix~A.2).

	{\flushleft \textbf{Multimodal Associative Memory}.}
	We use a multimodal memory for storing visual ($v$) and sentence ($s$) features with each modality while preserving the correspondence between them~\cite{working_memory}. 
	Information organization is more structured than the single modality memory as used in~\citet{MANN} and cross modality retrieval is straightforward under this design.
	A visual encoder implemented as a convolutional neural network followed by fully connected layers is used to encode the visual image $\mathbf{v}$ into a visual key $\mathbf{k}_{v}$, and then the corresponding sentence feature can be retrieved from the memory as:
	\begin{equation}
	\label{eq:read}
	\mathbf{r} \leftarrow \textbf{\textsc{Read}}(\mathbf{k}_{v},  \mathbf{M}_v, \mathbf{M}_s).
	\end{equation}
	$\mathbf{M}_v$  and $\mathbf{M}_s$  are  memories for visual  and sentence modalities  with the same number of  slots (columns).
	Memory read is implemented as  $\mathbf{r}\!\!=\!\!\mathbf{M}_s\boldsymbol\alpha$  with $\boldsymbol{\alpha}$ a soft reading weight  obtained through the visual modality by calculating the cosine similarities between $\mathbf{k}_{v}$ and  slots of $\mathbf{M}_v$.

	Memory write is similar to Neural Turing Machine~\cite{NTM}, but with a content importance gate $g_{\rm mem}$ to  adaptively control whether the content $\mathbf{c}$ should be written into  memory:
	\begin{equation}
	\label{eq:write}
	\nonumber
	\mathbf{M}_{m} \leftarrow \textbf{\textsc{Write}}(\mathbf{M}_m, \mathbf{c}_{m},  g_{\rm mem}),\; m\!\in\!\{v, s\}.
	\end{equation}
	For the visual modality $\mathbf{c}_v\!\! \triangleq \!\!\mathbf{k}_v$. 
	For the sentence modality,  
	$\mathbf{c}_s$  has to be selectively extracted from the sentence generated by the teacher.
	We use an attention mechanism to achieve this  by $\mathbf{c}_s\!\!\!=\!\!\!\mathbf{W} \mathbf{\boldsymbol{\eta}}$,
	where $\mathbf{W}$ denotes the matrix with columns being the embedding vectors of all the words in the sentence. 
	$\boldsymbol{\eta}$~is a normalized attention vector representing the relative importance of each word in the sentence as measured by the cosine similarity between the  sentence representation vector and each word's context vector, computed using a bidirectional-RNN.  
	The scalar-valued content importance gate $g_{\rm mem}$ is computed as a function of the sentence from the teacher,
	meaning that the importance of the content to be written into memory depends on the content itself (\emph{c.f.}, Appendix~A.3 for more details).
	The memory write  is achieved with an erase and an add operation:
	\begin{eqnarray}
	\label{eq:erase_add}
	\nonumber
	\begin{split}
	\; \scalebox{1}{{$\tilde{\mathbf{M}}_m$}} & =  \scalebox{1}{$\mathbf{M}_m -\mathbf{M}_m \odot ( g_{\rm mem} \cdot \mathbf{1} \cdot \boldsymbol{\beta}^\mathsf{T})$}, \\
	\; \scalebox{1}{$\mathbf{M}_{m}$} &= \scalebox{1}{$\tilde{\mathbf{M}}_m +   g_{\rm mem} \cdot  \mathbf{c}_m \cdot \boldsymbol{\beta}^\mathsf{T}$}, \; m\!\!\in\!\!\{v, s\}.
	\end{split}
	\end{eqnarray}
	$\odot$ denotes Hadamard product and the write location $\boldsymbol{\beta}$ is determined with a Least Recently Used
	Access mechanism~\cite{MANN}.

	\subsection{Context-adaptive Behavior Shaping through Reinforcement Learning}
	\label{RL_hRNN}
	
	Imitation fosters  the basic language ability for generating echoic behavior~\cite{BFSkinner}, but it is not enough for conversing adaptively with the teacher according to context and the knowledge state of the learner.  
	Thus we leverage reward feedback to shape the behavior of the agent by optimizing the policy using RL.
	The agent's response $\mathbf{a}^t$ is generated by
	the \emph{speaker}, which can  be modeled as a sample from a distribution over all possible  sequences, given the conversation history $\mathcal{H}^t\!\!=\!\!\{\mathbf{w}^{1}, \mathbf{a}^{1}, \cdots, \mathbf{w}^{t}\}$ and visual input $\mathbf{v}^t$:
	\begin{equation}
	\label{eq:action_generation}
	\mathbf{a}^t \sim p_{\theta}^{\rm S}(\mathbf{a}|\mathcal{H}^t, \mathbf{v}^{t}).
	\end{equation}
	As  $\mathcal{H}^t$ can be encoded by the interpreter-RNN as $\mathbf{h}^t_{\rm I}$, the action policy  can be represented as $p_{\theta}^{\rm S}(\mathbf{a}|\mathbf{h}^t_{\rm I}, \mathbf{v}^t)$.
	To leverage the language skill that is learned via imitation through the \emph{interpreter}, we can generate the sentence by implementing the \emph{speaker} with an RNN, sharing parameters with the interpreter-RNN, but with a conditional signal modulated by a controller network  (Figure~\ref{fig:network_structure}):
	\begin{eqnarray}
	\label{eq:action_prob}
	\begin{split}
	p_{\theta}^{\rm S}(\mathbf{a}^{t}|\mathbf{h}^t_{\rm I}, \mathbf{v}^t) = p_{\theta}^{\rm I}(\mathbf{a}^{t}|\mathbf{h}^t_{\rm I}+f(\mathbf{h}^t_{\rm I}, c), \mathbf{v}^t).
	\end{split}
	\end{eqnarray}
	The reason for using a controller $f(\cdot)$ for modulation is that 
	the basic language model only offers the learner the echoic ability to generate a sentence, but not necessarily the  adaptive behavior according to context (\emph{e.g.}   asking questions when facing novel objects and providing an answer for a previously learned object according to its own knowledge state).
	Without any additional module or learning signals, the agent's behaviors would be the same as those of the teacher because of parameter sharing; thus, it is difficult for the agent  to learn to  speak  in an adaptive manner. 
	
	To learn from consequences of speaking actions,
	the policy $p_{\theta}^{\rm S}(\cdot)$ is adjusted by maximizing expected future reward as represented by $\mathcal{L}^{\rm R}_{\theta}$. 
	As a non-differentiable sampling operation is involved  in Eqn.(\ref{eq:action_generation}), policy gradient theorem~\cite{Sutton}
	is used to derive the  gradient for updating $p_{\mathbf{\theta}}^{\rm S}(\cdot)$ in the reinforce module: 
	\begin{eqnarray}
	\label{policy_gradient}
	\begin{split}
	\nabla_{\theta} \mathcal{L}^{\rm R}_{\theta}  = \mathbb{E}_{p^{\rm S}_{\theta}}\big[ {\textstyle\sum_t} A^t\cdot \nabla_{\theta} \log p_{\mathbf{\theta}}^{\rm S}(\mathbf{a}^t | \mathbf{c}^t) \big],
	\end{split}
	\end{eqnarray}
	where  $A^t\!\!=\!\! V(\mathbf{h}^{t}_{\rm I}, c^t) \!-\! r^{t+1} \!-\! \gamma V(\mathbf{h}^{t+1}_{\rm I}, c^{t+1})$ is the advantage~\cite{Sutton} estimated using a value network $V(\cdot)$.
	The imitation module contributes by implementing $\mathcal{L}^{\rm I}_{\theta}$ with a cross-entropy loss~\cite{RL_Seq_ICLR15} and minimizing it with respect to  the parameters in $p_{\theta}^{\rm I}(\cdot)$, which are shared with $p_{\theta}^{\rm S}(\cdot)$.   
	The  training signal from imitation takes the shortcut connection without going through the controller. More details  on $f(\cdot)$, $V(\cdot)$ are provided in Appendix A.2.

	\section{Experiments}
	\label{sec:Exp}
	
	We conduct experiments with comparison to baseline approaches. 
	We first experiment with a word-level task in which the teacher and the learner communicate a single word each time. 
	We then investigate the impact of image variations on  concept learning. 
	We further  perform evaluation on the  more  challenging  sentence-level task in which the teacher and the agent communicate  in the form of sentences with varying lengths.

	\begin{figure}[t]
		\centering
		\begin{overpic}[viewport = 20 10 550 390, clip, width = 8cm]{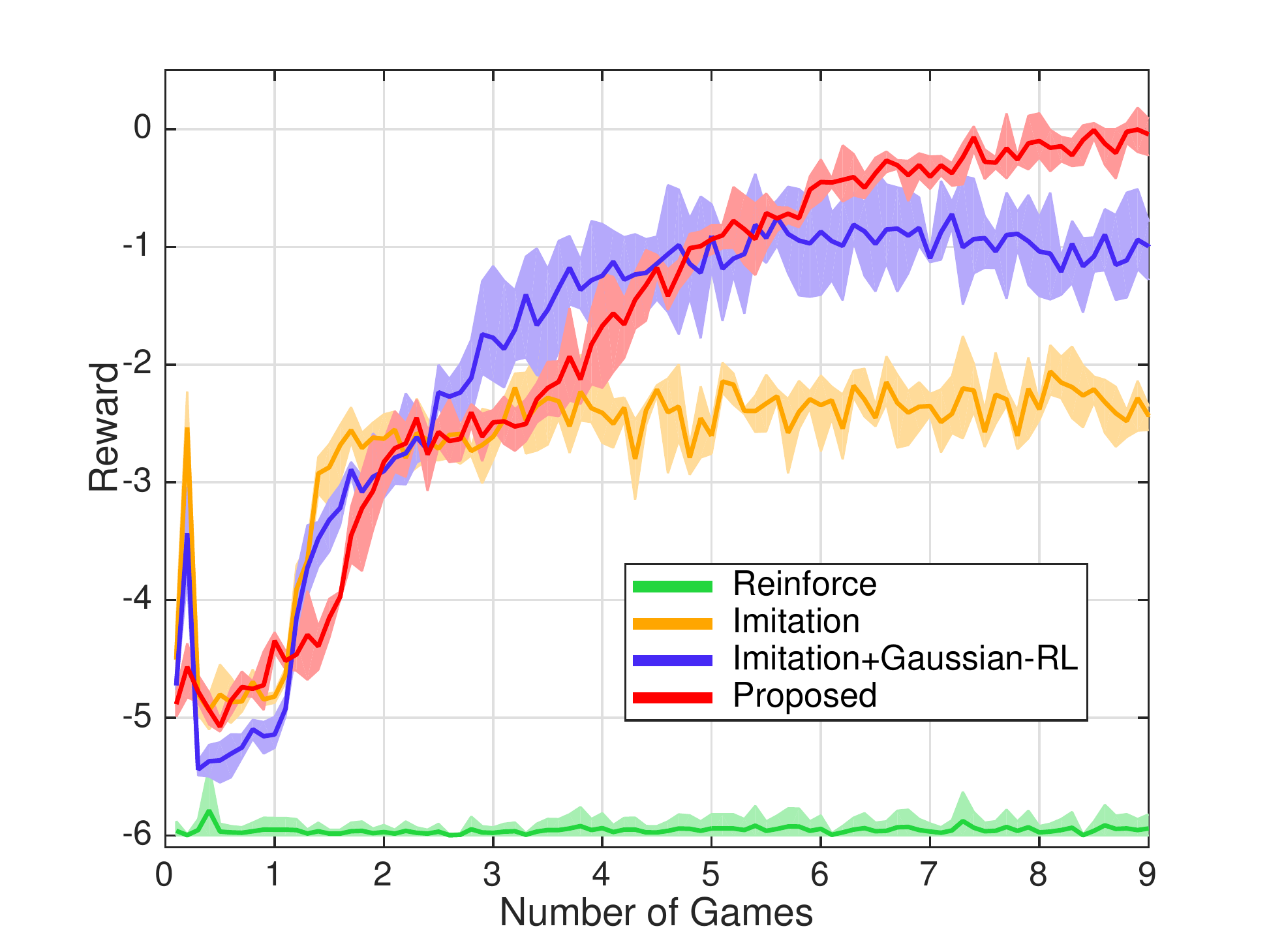}
			\put(88,0){\sffamily \scriptsize {\textcolor{black}{{$\times 10^3$}}}}
		\end{overpic}
		\vspace{-0.25in}
		\caption{\textbf{Evolution of reward} during training for the word-level task without image variations.} 
		\label{fig:traing_curve_word}	
	\end{figure}

	{\flushleft \textbf{Setup}}. 
	To evaluate the performance in learning a transferable  ability, rather than the ability of fitting a particular dataset, we use an {\small\textsf{Animal}} dataset for training and test the trained models on a {\small\textsf{Fruit}} dataset (Figure~\ref{fig:setting}).
	More details on the datasets are provided in Appendix~A.1.
	Each session consists of two randomly sampled classes, and the maximum number of interaction steps is six.
	
	{\flushleft \textbf{Baselines}.} The following methods are compared:
	\vspace{-0.05in}
	\begin{itemize}[leftmargin=10pt]
		\itemsep-0.3em
		\item \textbf{Reinforce}: a baseline model with the same network structure as the proposed model and trained using RL only, \emph{i.e.} minimizing $\mathcal{L}^{\rm R}_{\theta}$;
		\item \textbf{Imitation}: a  recurrent encoder decoder~\cite{HRNN_dialogue16}  model  with the same structure  as ours and trained via  imitation (minimizing~$\mathcal{L}^{\rm I}_{\theta}$);
		\item \textbf{Imitation+Gaussian-RL}: a joint imitation and reinforcement  method  using  a Gaussian policy~\cite{Gaussian_policy} in the latent  space of the control vector $\mathbf{c}^t$~\cite{L2T}.  
		The policy is changed by modifying the control vector $\mathbf{c}^t$ the action policy depends upon.
	\end{itemize}	
	
	{\flushleft \textbf{Training Details}.}
	The training algorithm is implemented with the deep learning platform {\fontsize{8}{12}\selectfont\sffamily PaddlePaddle}.\footnote{\fontsize{8.2}{12}\selectfont \url{https://github.com/PaddlePaddle/Paddle}}
	The whole network is trained from scratch in an end-to-end fashion. The network is randomly initialized without any pre-training and is trained with decayed 
	Adagrad~\cite{Adagrad}. We use a batch size of $16$, a learning rate of $1\!\!\times\!\!10^{-5}$ and a weight decay rate of $1.6\!\times \!10^{-3}$.
	We also exploit experience replay~\cite{AC_Replay, Nav}. 
	The reward discount factor $\gamma$ is $0.99$, the word embedding dimension $d$ is $1024$ and  the dictionary size $k$ is $80$.
	The visual image size is $32\!\times\!32$, the maximum length of generated sentence is $6$ and the memory size is $10$.
	Word embedding vectors are initialized as random vectors and remain fixed during training.
	A sampling operation is used for sentence generation during training for exploration while a $\max$ operation is used during testing both for \textbf{Proposed} and  for \textbf{Reinforce} baseline.
	The $\max$ operation is used in both training and testing for \textbf{Imitation} and \textbf{Imitation+Gaussian-RL} baselines.

	\subsection{Word-Level Task}
	In this experiment, we focus on a word-level task, which offers an opportunity to analyze and understand the underlying behavior of different algorithms while being free from distracting factors.
	Note that although the teacher speaks a word each time, the learner still has to learn to generate a full-sentence ended with an end-of-sentence symbol.

	\begin{figure}[t]
		\begin{overpic}[viewport = 14 45 544 405 clip, width = 3.8cm]{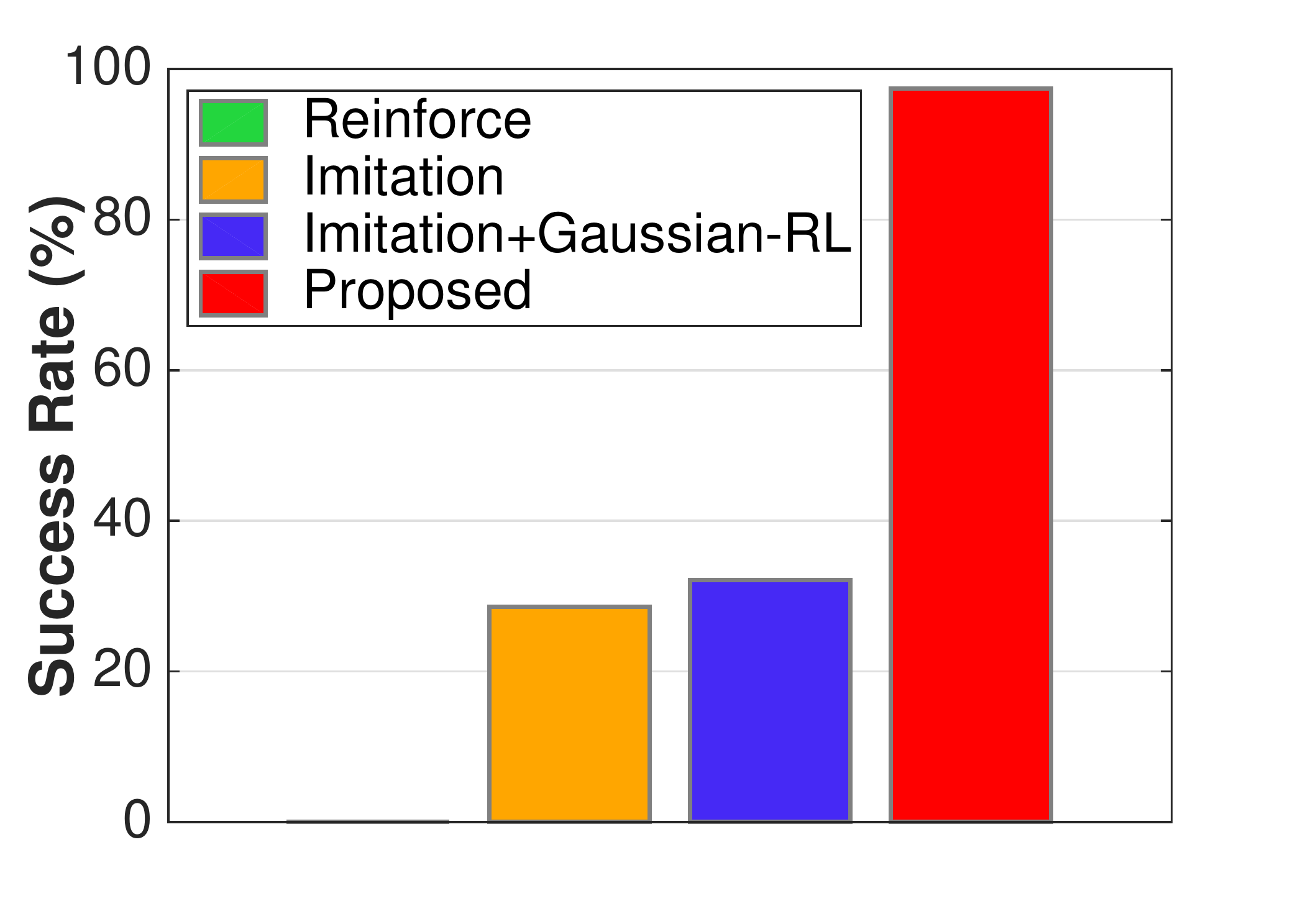}
		\end{overpic}
		\begin{overpic}[viewport = 20 45 550 405, clip, width = 3.8cm]{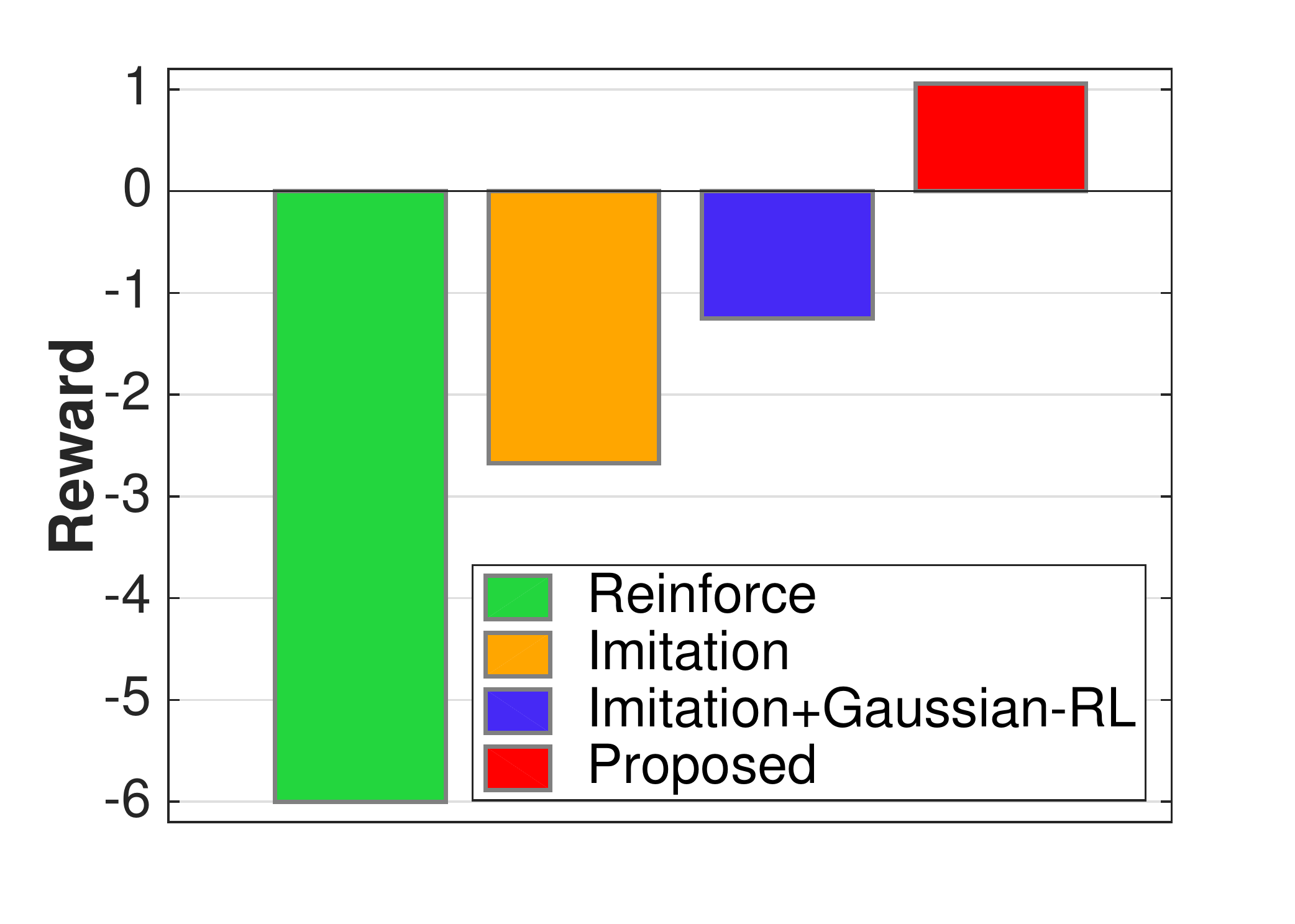}
		\end{overpic}
		\vspace{-0.2in}
		\caption{\textbf{Test performance} for the word-level task without image variations.
			Models are trained on the {\footnotesize\textsf{Animal}} dataset and tested on the {\footnotesize \textsf{Fruit}} dataset.} 
		\label{fig:word_no_var}	
		\vspace{-0.01in}	
	\end{figure}
	
	\begin{figure}[t]
		\begin{overpic}[viewport = 10 0 520 400, clip, width = 3.8cm]{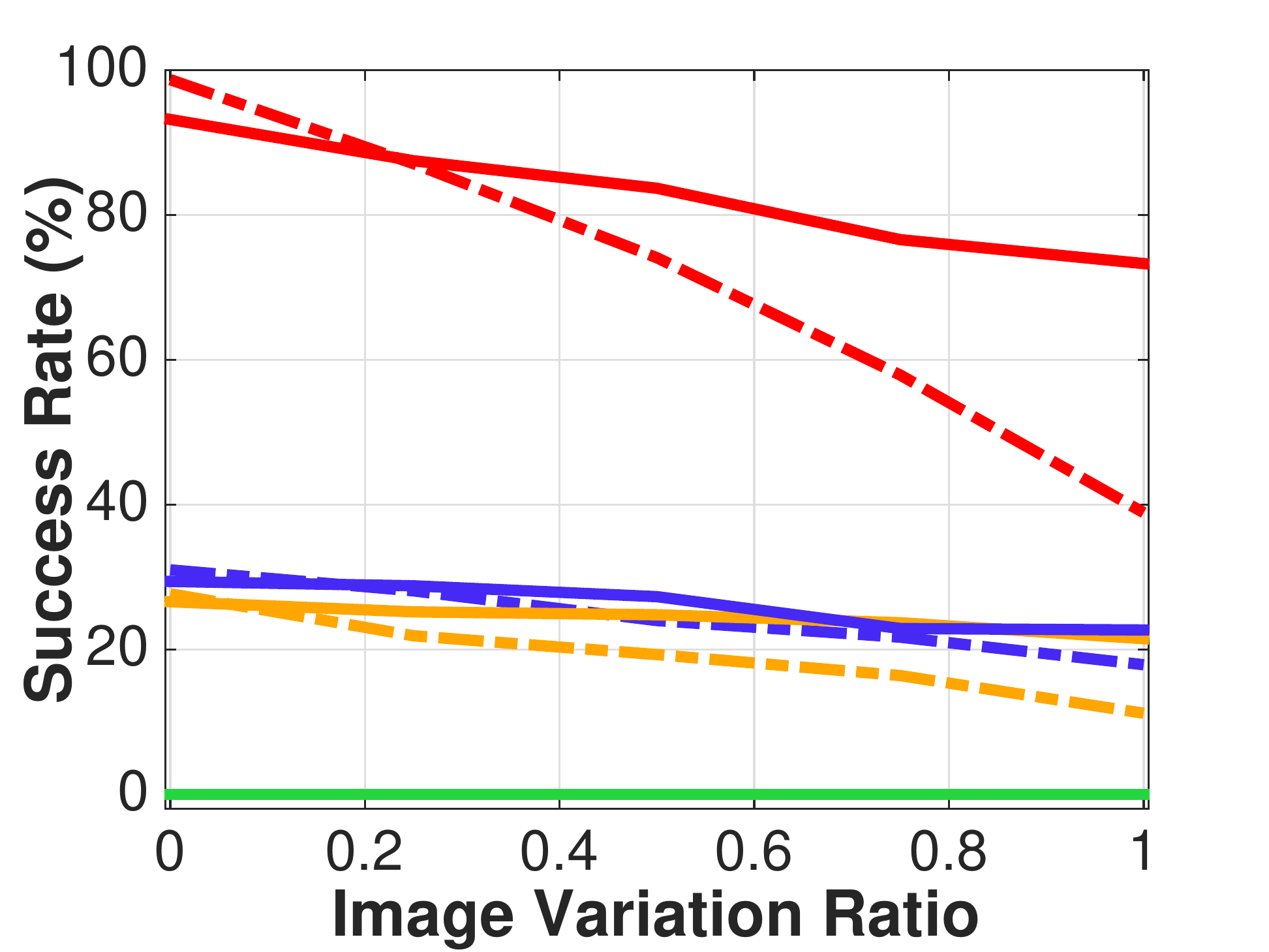}
		\end{overpic}
		\begin{overpic}[viewport = 20 0 530 400, clip,  width = 3.8cm]{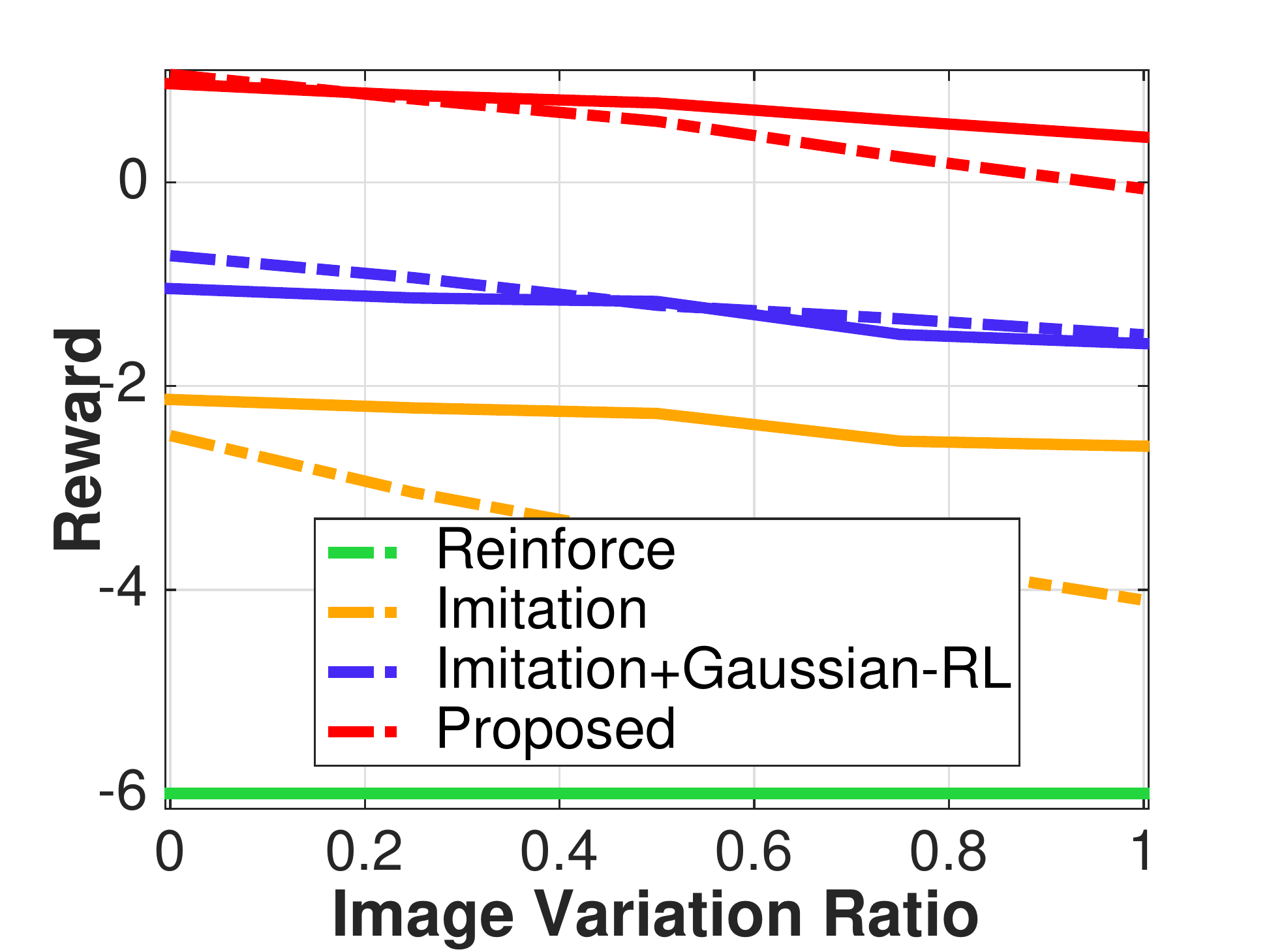}
		\end{overpic}
		\vspace{-0.2in}	
		\caption{\textbf{Test success rate and reward} for the word-level task on the {\footnotesize \textsf{Fruit}} dataset under different test image variation ratios for models trained on the {\footnotesize\textsf{Animal}} dataset with a variation ratio of 0.5 (solid lines) and without variation (dashed lines).} 
		\label{fig:word_with_var}
		\vspace{-0.1in}		
	\end{figure}

		\begin{figure*}[h]
			\centering
			\begin{overpic}[viewport = 110 30 450 390, clip, height = 4.1cm]{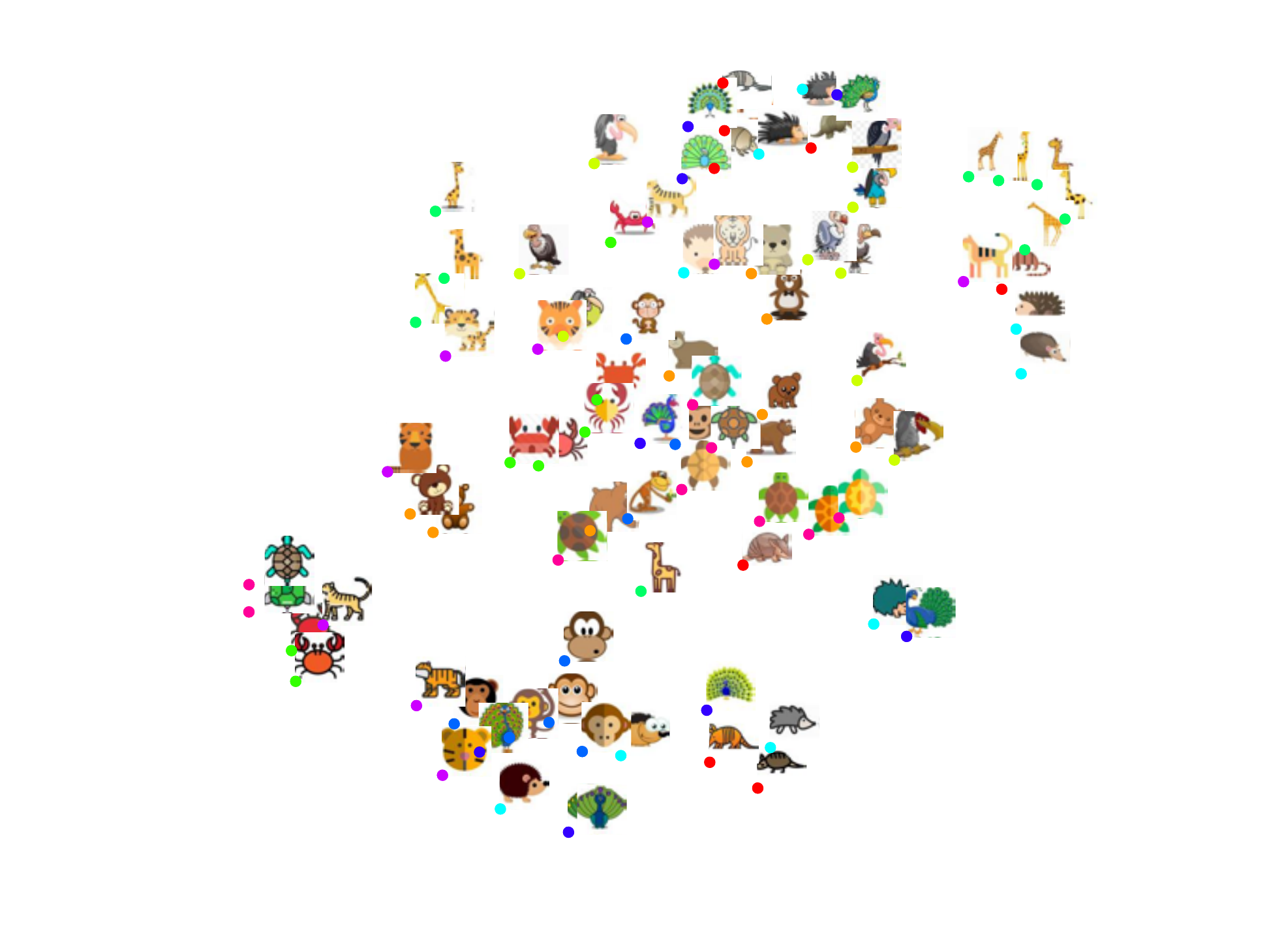}
				\put(10,5){\sffamily \scriptsize {\textcolor{black}{{(a)}}}}
			\end{overpic}
			\quad\;
			\begin{overpic}[viewport = 110 30 450 400, clip, height = 4.1cm]{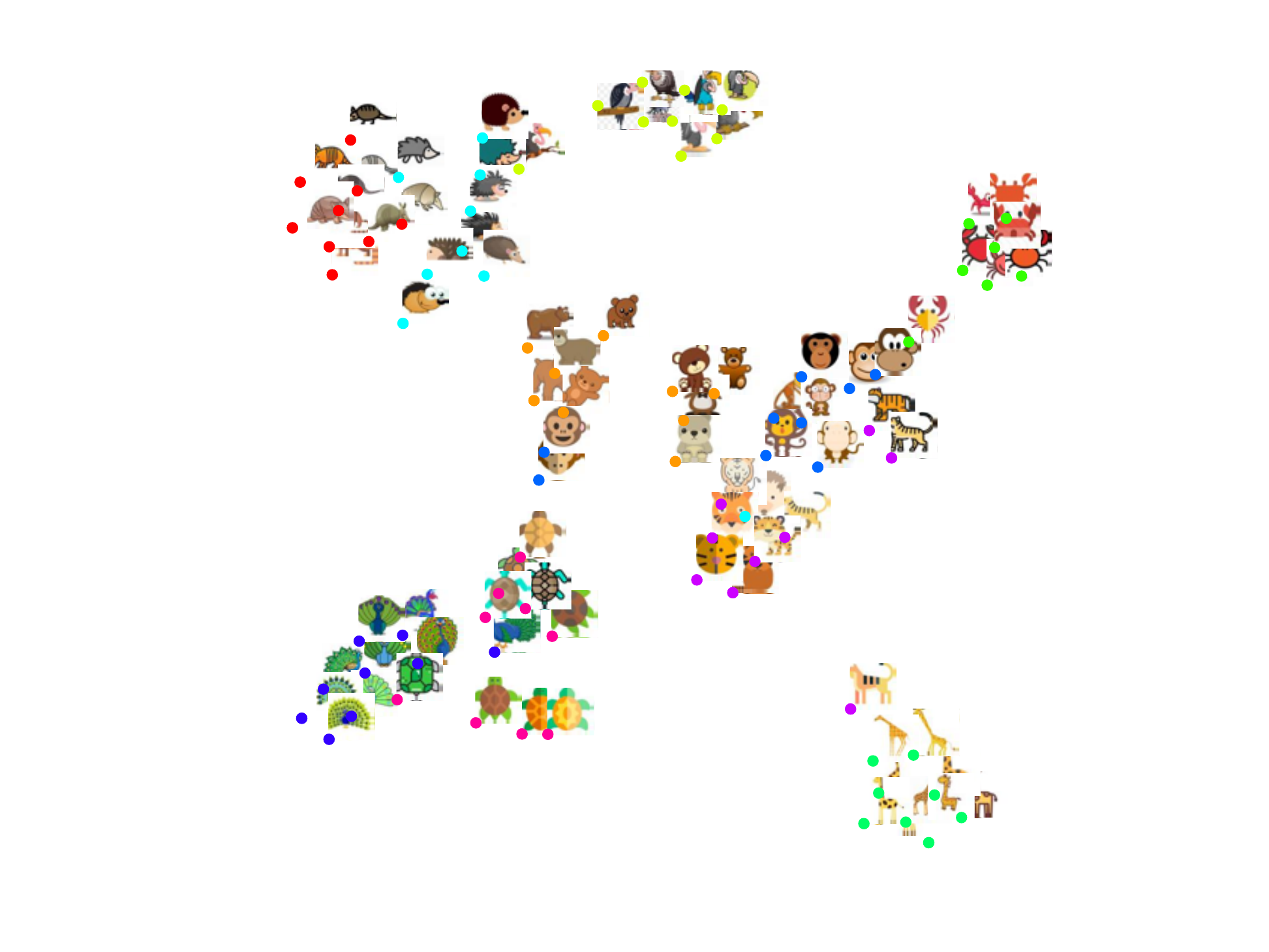}
				\put(10,5){\sffamily \scriptsize {\textcolor{black}{{(b)}}}}
			\end{overpic}
			\quad
			\begin{overpic}[viewport = 130 30 450 400, clip, height = 4.1cm]{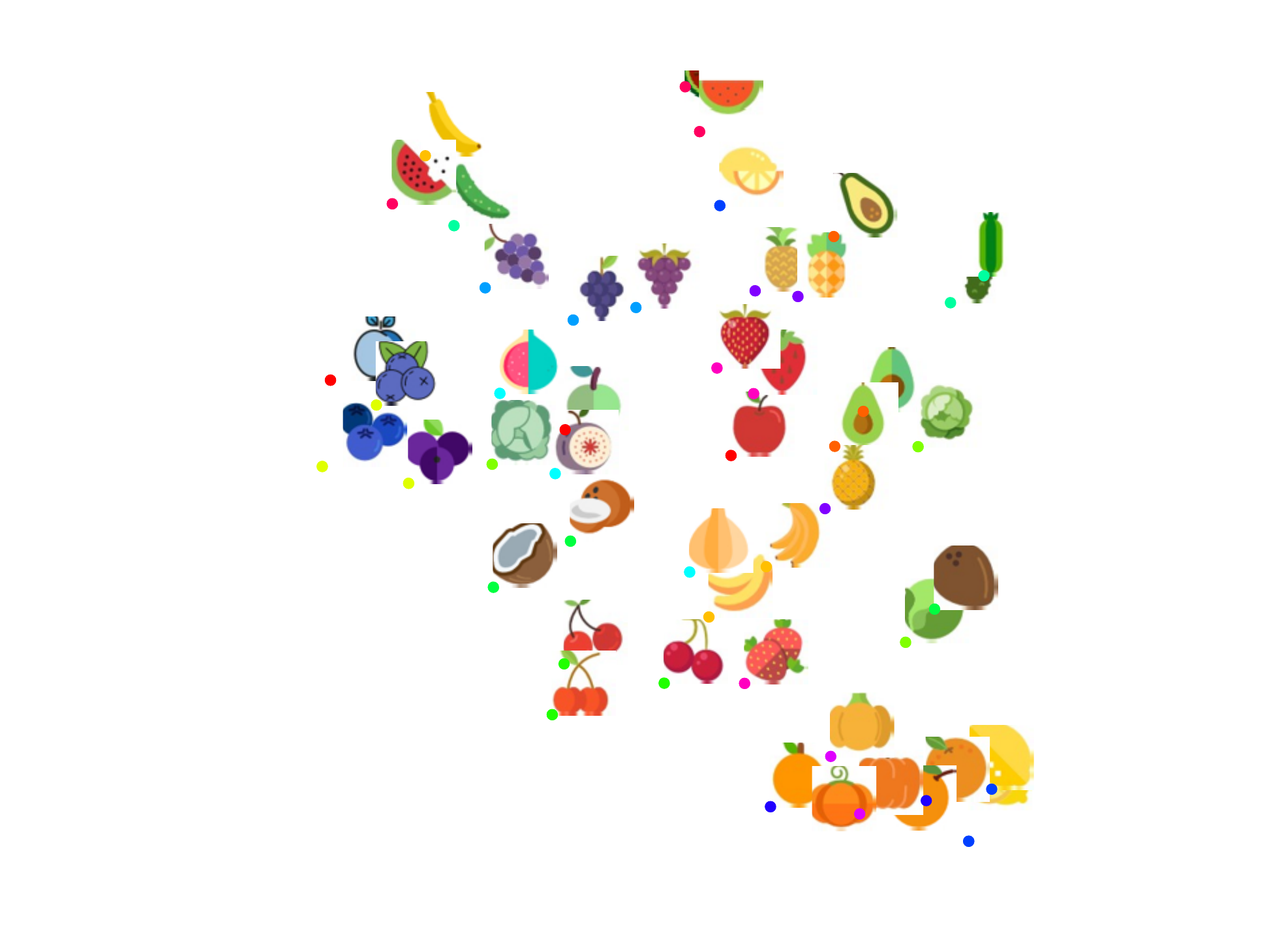}
				\put(10,5){\sffamily \scriptsize {\textcolor{black}{{(c)}}}}
			\end{overpic}
			\hspace{0.09in}
			\begin{overpic}[viewport = 140 30 450 400, clip, height = 4cm]{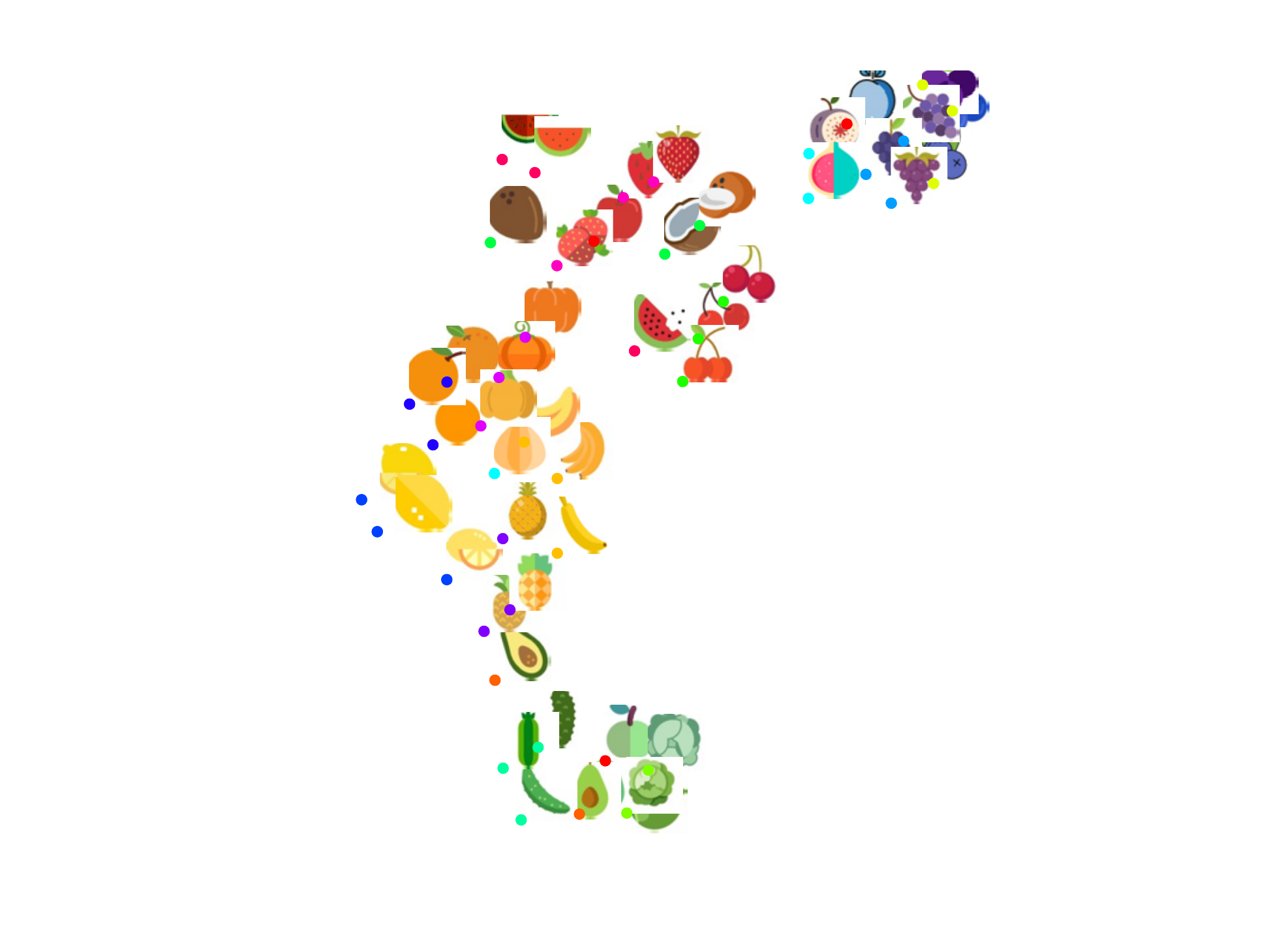}	
				\put(10,5){\sffamily \scriptsize {\textcolor{black}{{(d)}}}}	
			\end{overpic}
			\vspace{-0.2in}	
			\caption{\textbf{Visualization of the CNN features} with t-SNE.  Ten classes randomly sampled  from \textbf{(a-b)}~the {\footnotesize\textsf{Animal}} dataset and \textbf{(c-d)}~the {\footnotesize\textsf{Fruit}} dataset,
				with features extracted using the visual encoder trained without (a, c) and  with (b, d) image variations on the the {\footnotesize\textsf{Animal}} dataset.}
			\label{fig:cnn_viz}
			\vspace{-0.1in}
		\end{figure*}

		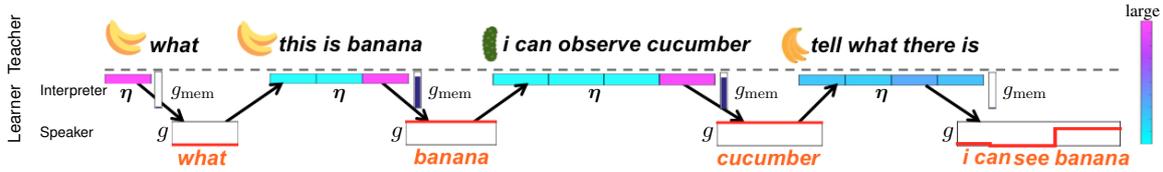
\begin{figure*}[t]
			\centering
			\hspace{0.4in}
			\begin{overpic}[viewport = 0 5 1150 190, clip,width=14cm]{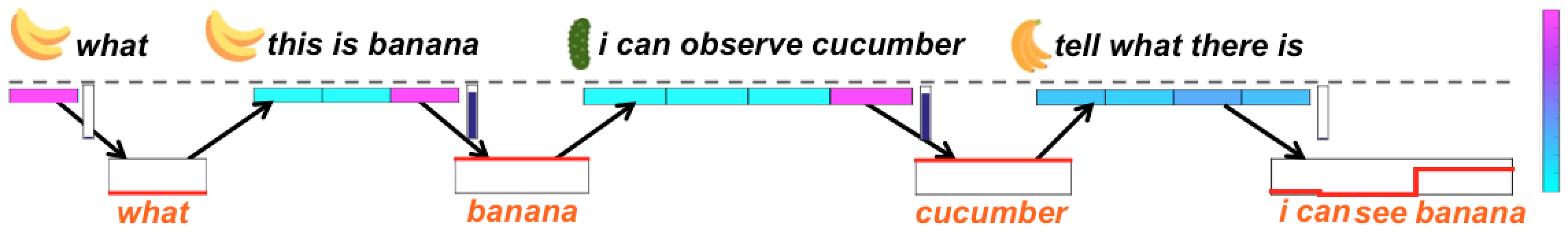}
				\put(-8.5,9.5){\sffamily \small {\textcolor{black}{{\scalebox{.7}{\rotatebox{90}{Teacher}}}}}}		
				\put(-8.5,3){\sffamily \small {\textcolor{black}{{\scalebox{.7}{\rotatebox{90}{Learner}}}}}}
				\put(-5.5,7.5){\sffamily \small {\textcolor{black}{{\scalebox{.6}{\rotatebox{0}{Interpreter}}}}}}
				\put(-5.5,3.5){\sffamily \small {\textcolor{black}{{\scalebox{.6}{\rotatebox{0}{Speaker}}}}}}
				\put(2,7.2){{{\textcolor{black}{\scalebox{.7}{$\boldsymbol{\eta}$}}}}}		
				\put(22,7.2){{{\textcolor{black}{\scalebox{.7}{$\boldsymbol{\eta}$}}}}}
				\put(46,7.2){{{\textcolor{black}{\scalebox{.7}{$\boldsymbol{\eta}$}}}}}
				\put(73,7.2){{{\textcolor{black}{\scalebox{.7}{$\boldsymbol{\eta}$}}}}}
				\put(5.5,3.5){{\colorbox{white}{\textcolor{black}{\scalebox{.8}{$g$}}}}}
				\put(27.5,3.5){{\colorbox{white}{\textcolor{black}{\scalebox{.8}{$g$}}}}}
				\put(56.7,3.5){{\colorbox{white}{\textcolor{black}{\scalebox{.8}{$g$}}}}}
				\put(79.3,3.5){{\colorbox{white}{\textcolor{black}{\scalebox{.8}{$g$}}}}}
				\put(6.8, 7.5){{{\textcolor{black}{\scalebox{.7}{$g_{\rm mem}$}}}}}
				\put(31, 7.5){{{\textcolor{black}{\scalebox{.7}{$g_{\rm mem}$}}}}}
				\put(59.8, 7.5){{{\textcolor{black}{\scalebox{.7}{$g_{\rm mem}$}}}}}
				\put(85, 7.5){{{\textcolor{black}{\scalebox{.7}{$g_{\rm mem}$}}}}}
				\put(96.5,15){{{\textcolor{black}{\scalebox{.6}{large}}}}}
			\end{overpic}
			\vspace{-0.1in}	
			\caption{\textbf{Example  results} of the proposed approach on novel classes. The learner can ask about the new class and use the interpreter  to extract useful information  from the teacher's sentence via word-level attention
				$\boldsymbol{\eta}$ and content importance $g_{\rm mem}$ jointly.
				The speaker uses the fusion gate $g$ to adaptively switch between signals from RNN (small $g$) and external memory (large $g$) to generate sentence responses.}  
			\label{fig:sent_with_attention}	
			\vspace{-0.05in}		
		\end{figure*}
		
	Figure~\ref{fig:traing_curve_word} shows the evolution curves of the rewards during training for different approaches. It is observed  that \textbf{Reinforce}  makes very little progress, mainly due to the difficulty of exploration in the large   space of sequence actions.
	\textbf{Imitation} obtains higher rewards than \textbf{Reinforce}  during training, as it can avoid some penalty by generating  sensible sentences such as questions.
	\textbf{Imitation+Gaussian-RL} gets higher rewards than both \textbf{Imitation} and \textbf{Reinforce}, indicating that the RL component reshapes the action policy toward higher rewards. However, as the Gaussian policy optimizes the action policy indirectly in a latent feature space, it is less efficient for exploration and learning.
	\textbf{Proposed} achieves the highest final reward during training.

	We train the models  using the {\small\textsf{Animal}} dataset and  evaluate them on the {\small\textsf{Fruit}}  dataset; Figure~\ref{fig:word_no_var} summarizes  the success rate and average reward over 1K testing sessions. 
	As can be observed, \textbf{Reinforce}  achieves the lowest success rate ($0.0\%$) and reward ($-6.0$) due to its inherent inefficiency in learning.
	\textbf{Imitation}  performs better than \textbf{Reinforce}  in terms of both its success rate ($28.6\%$) and reward value ($-2.7$). \textbf{Imitation+Gaussian-RL} achieves  a higher reward ($-1.2$) during testing, but its success rate ($32.1\%$) is similar to that of 
	 \textbf{Imitation}, mainly due to the rigorous criteria for success.  \textbf{Proposed} reaches the highest success rate ($97.4\%$) and average reward ($+1.1$)\footnote{The testing reward is higher than the training reward mainly due to the action sampling in training for exploration. }, outperforming all baseline methods by a large margin.
	From this experiment, it is clear that imitation with a proper usage of reinforcement is crucial for achieving  adaptive  behaviors (\emph{e.g.}, asking questions about novel objects and generating answers or statements  about learned objects proactively).

	\subsection{Learning with Image Variations}
	To evaluate the impact of within-class image variations on one-shot concept learning,
	we train models with and without image variations, and during testing compare their performance under different image variation ratios  (the chance  of a novel image instance  being present within a session)， as shown in Figure~\ref{fig:word_with_var}.
	It is observed that the performance of the model trained without image variations drops significantly as the variation ratio increases.
	We also evaluate the performance of models trained under a variation ratio of $0.5$.
	Figure~\ref{fig:word_with_var} clearly shows that although there is also a performance drop, which is expected, the performance degrades more gradually, indicating the  importance of image variation for learning one-shot concepts.
	Figure~\ref{fig:cnn_viz} visualizes sampled training and testing images represented by their corresponding features extracted using the visual encoder trained without and with image variations.
	Clusters of visually similar concepts emerge in the feature space when trained with image variations, indicating that a more discriminative visual encoder was obtained for learning generalizable concepts.

	\subsection{Sentence-Level Task}
\begin{figure}[t]
		\begin{overpic}[viewport = 14 45 544 405 clip, width = 3.8cm]{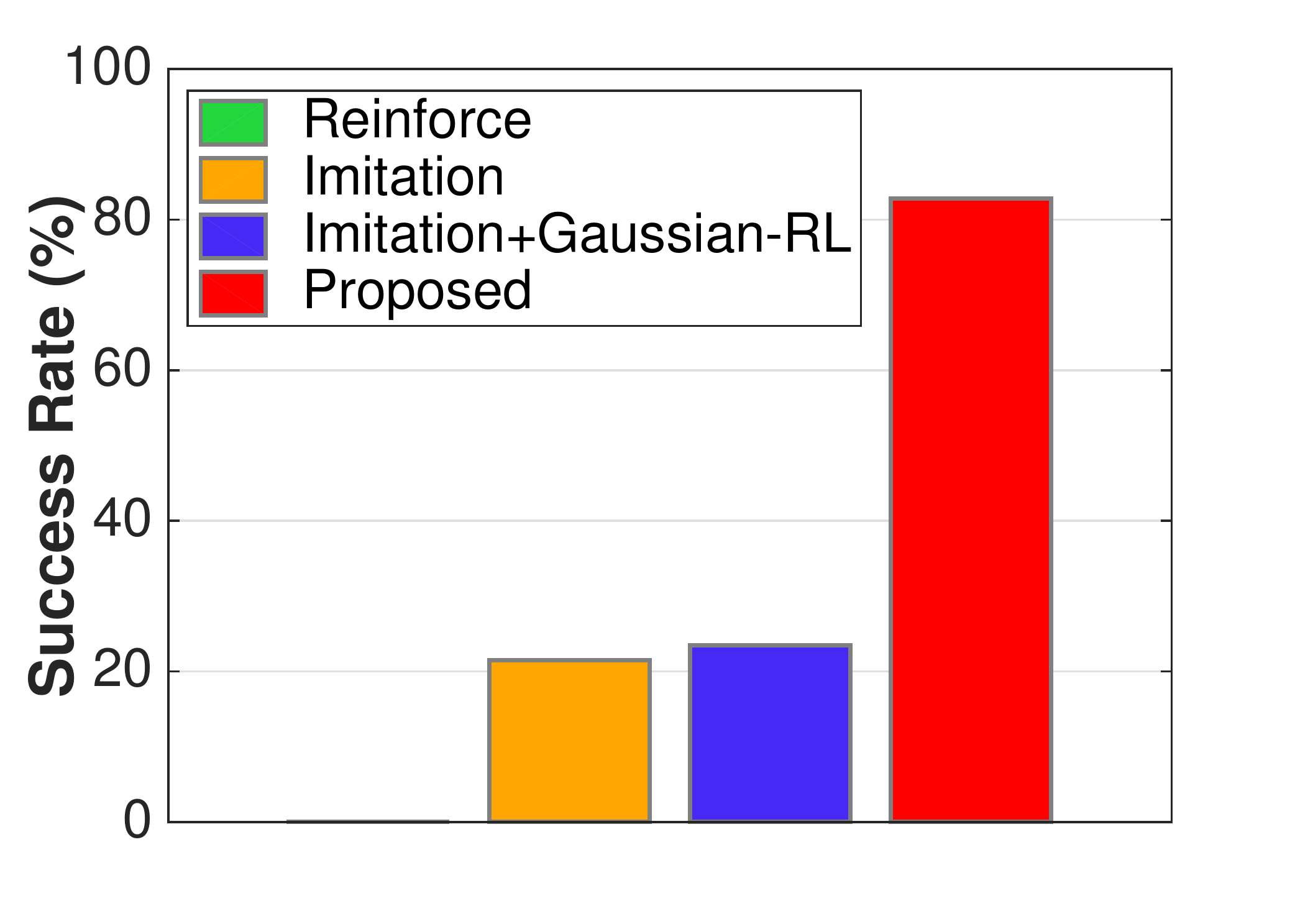}
		\end{overpic}
		\begin{overpic}[viewport = 20 45 550 405, clip, width = 3.8cm]{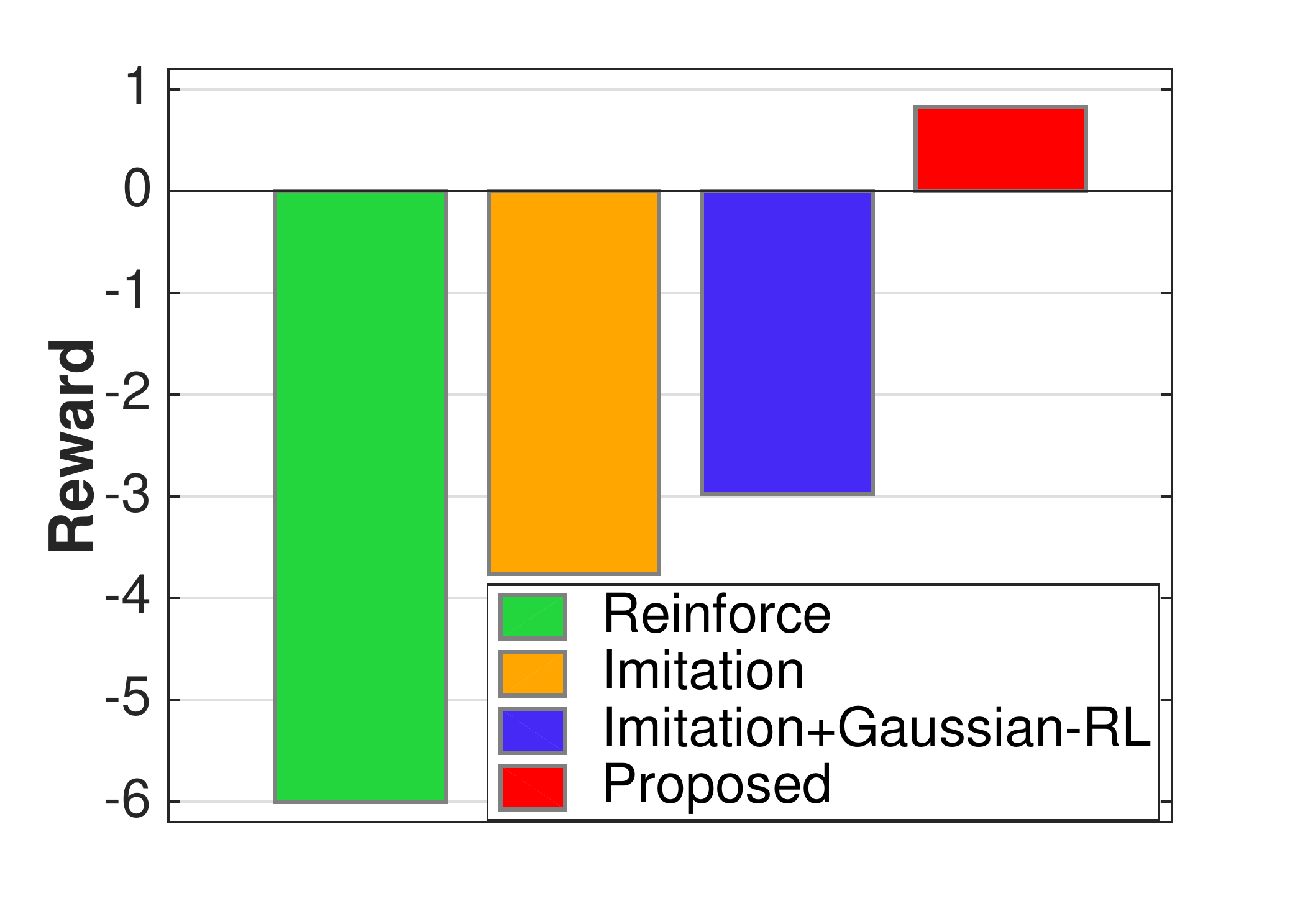}
		\end{overpic}
		\caption{\textbf{Test performance} for sentence-level task with image variations (variation ratio=0.5).} 
		\label{fig:sent_var}
	\end{figure}

	We further evaluate the model on  sentence-level tasks.
	Teacher's sentences are generated using the grammar as shown in Table~\ref{tab:grammar} and have a number of variations  with sentence lengths ranging from one to five.
	Example sentences from the teacher are presented in Appendix~A.1.
	This task is more challenging than the word-level task in two ways: i) information processing is more difficult as the learner has to learn to extract useful information which could appear at different locations of the sentence;
	ii) the sentence generation is also more difficult than the word-level task and the learner has to adaptively fuse information from RNN and external memory to generate a complete sentence.
	Comparison of different approaches in terms of their success rates  and average rewards  on the novel test set are shown in Figure~\ref{fig:sent_var}.
	As can be observed from the figure, \textbf{Proposed} again outperforms all  other compared methods  in terms of both success rate ($82.8\%$) and average reward (${\displaystyle+}0.8$), demonstrating its effectiveness even for the more complex sentence-level task.
		
	We also visualize the information extraction and the adaptive sentence composing process of the proposed approach when applied to a test set. As shown in Figure~\ref{fig:sent_with_attention}, the agent learns to extract useful information from the teacher's sentence and use the content importance gate to control what content is written into the external memory. Concretely, sentences containing object names have a larger $g_{\rm mem}$ value, and the word corresponding to object name has a larger value in the attention vector $\boldsymbol{\eta}$ compared to other words in the sentence.
	The combined effect of $\boldsymbol{\eta}$  and $g_{\rm mem}$ suggests that words corresponding to object names  have higher likelihoods of being written into the external memory.
	The agent also successfully learns to use the external memory for storing the information extracted from the teacher's sentence,  to fuse it adaptively with the signal from the RNN (capturing the syntactic structure) and to generate a complete sentence with the new concept included.
	The value of the fusion gate $g$ is small when generating words like ``\emph{what},'', ``\emph{i},'' ``\emph{can},'' and ``\emph{see},'' meaning it mainly relies on the signal from the RNN  for generation (\emph{c.f.}, Eqn.(\ref{eq:fuse})  and Figure~\ref{fig:sent_with_attention}).  In contrast, when generating object names (\emph{e.g.}, ``\emph{banana},'' and  ``\emph{cucumber}''), the fusion gate $g$ has a large value, meaning that there is more emphasis  on the signal from the external memory.
	This experiment showed that the proposed approach is applicable to the more complex sentence-level task for language learning and one-shot learning. More interestingly, it learns an interpretable operational process, which can be easily understood. 
	More results including example dialogues from different approaches are presented in Appendix~A.4.
	
	\section{Discussion}
	\label{sec:con}
	We have presented an approach for grounded language acquisition with one-shot visual concept learning in this work. This is achieved by purely interacting with a teacher and learning from feedback arising naturally during interaction through joint imitation and reinforcement learning, with a memory augmented neural network.
	Experimental results show that the proposed approach is effective for language acquisition with one-shot visual concept learning across several different settings compared with several baseline approaches.
	
	In the current work, we have designed and used a computer game (synthetic task with synthetic language) for training the agent. This is mainly due to the fact that there is no existing dataset to the best of our knowledge that is adequate for developing our addressed interactive language learning and one-shot learning problem.
	For our current design, although it is an artificial game, there is a reasonable amount of variations both within and across sessions, e.g., the object classes to be learned within a session, the presentation order of the selected classes, the sentence patterns and image instances to be used etc. All these factors contribute to  the increased complexity of the learning task, making it non-trivial and already very challenging to existing approaches as shown by the experimental results.
	While offering flexibility in training, one downside of using a synthetic task is its limited amount of variation compared with real-world scenarios with natural languages.
	Although it might be non-trivial to extend the proposed approach to real natural language directly, we regard this work as an initial step towards this ultimate ambitious goal and our game might shed some light on designing more advanced games or performing real-world data collection. 	
	We plan to investigate the generalization and application of the proposed approach to more realistic environments with more diverse tasks in future work.
	
	\section*{Acknowledgments}
	We thank the reviewers and PC members for theirs efforts in helping improving the paper.
	We thank Xiaochen Lian and Xiao Chu for their discussions.

	\bibliography{ref}

\begin{thebibliography}{40}
\expandafter\ifx\csname natexlab\endcsname\relax\def\natexlab#1{#1}\fi

\bibitem[{Antol et~al.(2015)Antol, Agrawal, Lu, Mitchell, Batra, Zitnick, and
  Parikh}]{VQA}
Stanislaw Antol, Aishwarya Agrawal, Jiasen Lu, Margaret Mitchell, Dhruv Batra,
  C.~Lawrence Zitnick, and Devi Parikh. 2015.
\newblock {VQA}: {V}isual {Q}uestion {A}nswering.
\newblock In \emph{International Conference on Computer Vision (ICCV)}.

\bibitem[{Baddeley(1992)}]{working_memory}
Alan Baddeley. 1992.
\newblock Working memory.
\newblock \emph{Science}, 255(5044):556–--559.

\bibitem[{Bahdanau et~al.(2017)Bahdanau, Brakel, Xu, Goyal, Lowe, Pineau,
  Courville, and Bengio}]{AC_Seq}
Dzmitry Bahdanau, Philemon Brakel, Kelvin Xu, Anirudh Goyal, Ryan Lowe, Joelle
  Pineau, Aaron~C. Courville, and Yoshua Bengio. 2017.
\newblock An actor-critic algorithm for sequence prediction.
\newblock In \emph{International Conference on Learning Representations
  (ICLR)}.

\bibitem[{Baker et~al.(2002)Baker, Jensen, and Kolb}]{conversational_learning}
Ann~C. Baker, Patricia~J. Jensen, and David~A. Kolb. 2002.
\newblock \emph{Conversational Learning: An Experiential Approach to Knowledge
  Creation}.
\newblock Copley Publishing Group.

\bibitem[{Borovsky et~al.(2003)Borovsky, Kutas, and Elman}]{oneshot_word}
Arielle Borovsky, Marta Kutas, and Jeff Elman. 2003.
\newblock Learning to use words: Event related potentials index single-shot
  contextual word learning.
\newblock \emph{Cognzition}, 116(2):289–--296.

\bibitem[{Cho et~al.(2014{\natexlab{a}})Cho, Merrienboer, Gülçehre, Bahdanau,
  Bougares, Schwenk, and Bengio}]{GRU}
K.~Cho, B.~Merrienboer, C.~Gülçehre, D.~Bahdanau, F.~Bougares, H.~Schwenk,
  and Y.~Bengio. 2014{\natexlab{a}}.
\newblock Learning phrase representations using rnn encoder-decoder for
  statistical machine translation.
\newblock In \emph{Empirical Methods in Natural Language Processing (EMNLP)}.

\bibitem[{Cho et~al.(2014{\natexlab{b}})Cho, van Merri{\"{e}}nboer,
  G{\"{u}}l{\c c}ehre, Bahdanau, Bougares, Schwenk, and
  Bengio}]{cho-al-emnlp14}
Kyunghyun Cho, Bart van Merri{\"{e}}nboer, {\c C}ağlar G{\"{u}}l{\c c}ehre,
  Dzmitry Bahdanau, Fethi Bougares, Holger Schwenk, and Yoshua Bengio.
  2014{\natexlab{b}}.
\newblock Learning phrase representations using {RNN} encoder--decoder for
  statistical machine translation.
\newblock In \emph{Empirical Methods in Natural Language Processing (EMNLP)}.

\bibitem[{Das et~al.(2017)Das, Kottur, , Moura, Lee, and Batra}]{visdial_rl}
Abhishek Das, Satwik Kottur, , Jos\'e~M.F. Moura, Stefan Lee, and Dhruv Batra.
  2017.
\newblock Learning cooperative visual dialog agents with deep reinforcement
  learning.
\newblock In \emph{International Conference on Computer Vision (ICCV)}.

\bibitem[{Duan et~al.(2016)Duan, Chen, Houthooft, Schulman, and
  Abbeel}]{Gaussian_policy}
Yan Duan, Xi~Chen, Rein Houthooft, John Schulman, and Pieter Abbeel. 2016.
\newblock Benchmarking deep reinforcement learning for continuous control.
\newblock In \emph{International Conference on International Conference on
  Machine Learning (ICML)}.

\bibitem[{Duchi et~al.(2011)Duchi, Hazan, and Singer}]{Adagrad}
J.~Duchi, E.~Hazan, and Y.~Singer. 2011.
\newblock Adaptive subgradient methods for online learning and stochastic
  optimization.
\newblock \emph{Journal of Machine Learning Research}, 12:2121--2159.

\bibitem[{Foerster et~al.(2016)Foerster, Assael, de~Freitas, and
  Whiteson}]{RL_Com}
Jakob~N. Foerster, Yannis~M. Assael, Nando de~Freitas, and Shimon Whiteson.
  2016.
\newblock Learning to communicate with deep multi-agent reinforcement learning.
\newblock In \emph{Advances in Neural Information Processing Systems (NIPS)}.

\bibitem[{Graves et~al.(2014)Graves, Wayne, and Danihelka}]{NTM}
Alex Graves, Greg Wayne, and Ivo Danihelka. 2014.
\newblock Neural turing machines.
\newblock \emph{CoRR}, abs/1410.5401.

\bibitem[{He et~al.(2016)He, Chen, He, Gao, Li, Deng, and
  Ostendorf}]{Lan_Action_Space}
Ji~He, Jianshu Chen, Xiaodong He, Jianfeng Gao, Lihong Li, Li~Deng, and Mari
  Ostendorf. 2016.
\newblock Deep reinforcement learning with a natural language action space.
\newblock In \emph{Association for Computational Linguistics (ACL)}.

\bibitem[{Houston and Miyamoto(2011)}]{hearing}
Derek~M. Houston and Richard~T. Miyamoto. 2011.
\newblock Effects of early auditory experience on word learning and speech
  perception in deaf children with cochlear implants: Implications for
  sensitive periods of language development.
\newblock \emph{Otol Neurotol}, 31(8):1248–--1253.

\bibitem[{Kuhl(2004)}]{nature_lang}
Patricia~K. Kuhl. 2004.
\newblock Early language acquisition: cracking the speech code.
\newblock \emph{Nat Rev Neurosci}, 5(2):831--843.

\bibitem[{Lake et~al.(2011)Lake, Salakhutdinov, Gross, and
  Tenenbaum}]{lake_oneshot}
Brenden~M. Lake, Ruslan Salakhutdinov, Jason Gross, and Joshua~B. Tenenbaum.
  2011.
\newblock One shot learning of simple visual concepts.
\newblock In \emph{Proceedings of the 33th Annual Meeting of the Cognitive
  Science Society}.

\bibitem[{Lazaridou et~al.(2017)Lazaridou, Peysakhovich, and
  Baroni}]{Multi_Agent_Lan}
Angeliki Lazaridou, Alexander Peysakhovich, and Marco Baroni. 2017.
\newblock Multi-agent cooperation and the emergence of (natural) language.
\newblock In \emph{International Conference on Learning Representations
  (ICLR)}.

\bibitem[{Li et~al.(2017)Li, Miller, Chopra, Ranzato, and
  Weston}]{dialogue_question}
Jiwei Li, Alexander~H. Miller, Sumit Chopra, Marc’Aurelio Ranzato, and Jason
  Weston. 2017.
\newblock Learning through dialogue interactions by asking questions.
\newblock In \emph{International Conference on Learning Representations
  (ICLR)}.

\bibitem[{Li et~al.(2016)Li, Monroe, Ritter, Jurafsky, Galley, and
  Gao}]{RL_Dialogue}
Jiwei Li, Will Monroe, Alan Ritter, Dan Jurafsky, Michel Galley, and Jianfeng
  Gao. 2016.
\newblock Deep reinforcement learning for dialogue generation.
\newblock In \emph{Empirical Methods in Natural Language Processing (EMNLP)}.

\bibitem[{Mnih et~al.(2013)Mnih, Kavukcuoglu, Silver, Graves, Antonoglou,
  Wierstra, and Riedmiller}]{DQN}
V.~Mnih, K.~Kavukcuoglu, D.~Silver, A.~Graves, I.~Antonoglou, D.~Wierstra, and
  M.~Riedmiller. 2013.
\newblock Playing {Atari} with deep reinforcement learning.
\newblock In \emph{NIPS Deep Learning Workshop}.

\bibitem[{Mordatch and Abbeel(2018)}]{Emergence_Lan}
Igor Mordatch and Pieter Abbeel. 2018.
\newblock Emergence of grounded compositional language in multi-agent
  populations.
\newblock In \emph{Association for the Advancement of Artificial Intelligence
  (AAAI)}.

\bibitem[{Petursdottir and Mellor(2016)}]{rl_lang_learning}
Anna~Ingeborg Petursdottir and James~R. Mellor. 2016.
\newblock Reinforcement contingencies in language acquisition.
\newblock \emph{Policy Insights from the Behavioral and Brain Sciences},
  4(1):25--32.

\bibitem[{Ranzato et~al.(2016)Ranzato, Chopra, Auli, and
  Zaremba}]{RL_Seq_ICLR15}
Marc'Aurelio Ranzato, Sumit Chopra, Michael Auli, and Wojciech Zaremba. 2016.
\newblock Sequence level training with recurrent neural networks.
\newblock In \emph{International Conference on Learning Representations
  (ICLR)}.

\bibitem[{Santoro et~al.(2016)Santoro, Bartunov, Botvinick, Wierstra, and
  Lillicrap}]{MANN}
Adam Santoro, Sergey Bartunov, Matthew Botvinick, Daan Wierstra, and Timothy
  Lillicrap. 2016.
\newblock Meta-learning with memory-augmented neural networks.
\newblock In \emph{International Conference on Machine Learning (ICML)}.

\bibitem[{Serban et~al.(2016)Serban, Sordoni, Bengio, Courville, and
  Pineau}]{HRNN_dialogue16}
Iulian~Vlad Serban, Alessandro Sordoni, Yoshua Bengio, Aaron~C. Courville, and
  Joelle Pineau. 2016.
\newblock Building end-to-end dialogue systems using generative hierarchical
  neural network models.
\newblock In \emph{Association for the Advancement of Artificial Intelligence
  (AAAI)}.

\bibitem[{Skinner(1957)}]{BFSkinner}
B.~F. Skinner. 1957.
\newblock \emph{Verbal Behavior}.
\newblock Copley Publishing Group.

\bibitem[{Stadie et~al.(2017)Stadie, Abbeel, and Sutskever}]{third_person}
Bradly~C. Stadie, Pieter Abbeel, and Ilya Sutskever. 2017.
\newblock Third-person imitation learning.
\newblock In \emph{International Conference on Learning Representations
  (ICLR)}.

\bibitem[{Stent and Bangalore(2014)}]{book_dialogue_supervised}
Amanda Stent and Srinivas Bangalore. 2014.
\newblock \emph{Natural Language Generation in Interactive Systems}.
\newblock Cambridge University Press.

\bibitem[{Strub et~al.(2017)Strub, de~Vries, Mary, Piot, Courville, and
  Pietquin}]{Dialogue_Deepmind}
Florian Strub, Harm de~Vries, J{\'{e}}r{\'{e}}mie Mary, Bilal Piot, Aaron~C.
  Courville, and Olivier Pietquin. 2017.
\newblock End-to-end optimization of goal-driven and visually grounded dialogue
  systems.
\newblock In \emph{International Joint Conference on Artificial Intelligence
  (IJCAI)}.

\bibitem[{Sukhbaatar et~al.(2016)Sukhbaatar, Szlam, and Fergus}]{BP_Com}
Sainbayar Sukhbaatar, Arthur Szlam, and Rob Fergus. 2016.
\newblock Learning multiagent communication with backpropagation.
\newblock In \emph{Advances in Neural Information Processing Systems (NIPS)}.

\bibitem[{Sutton and Barto(1998)}]{Sutton}
Richard~S. Sutton and Andrew~G. Barto. 1998.
\newblock \emph{Reinforcement Learning: An Introduction}.
\newblock MIT Press.

\bibitem[{Wang et~al.(2016)Wang, Liang, and Manning}]{wang2016games}
S.~I. Wang, P.~Liang, and C.~Manning. 2016.
\newblock Learning language games through interaction.
\newblock In \emph{Association for Computational Linguistics (ACL)}.

\bibitem[{Wang et~al.(2017)Wang, Bapst, Heess, Mnih, Munos, Kavukcuoglu, and
  Freitas}]{AC_Replay}
Z.~Wang, V.~Bapst, N.~Heess, V.~Mnih, R.~Munos, K.~Kavukcuoglu, and N.~Freitas.
  2017.
\newblock Sample efficient actor-critic with experience replay.
\newblock In \emph{International Conference on Learning Representations
  (ICLR)}.

\bibitem[{Waxman(2004)}]{game}
Sandra~R. Waxman. 2004.
\newblock \emph{Everything had a name, and each name gave birth to a new
  thought: links between early word learning and conceptual organization}.
\newblock Cambridge, MA: The MIT Press.

\bibitem[{Wen et~al.(2015)Wen, Gasic, Mrksic, Su, Vandyke, and
  Young}]{lang_generation}
Tsung{-}Hsien Wen, Milica Gasic, Nikola Mrksic, Pei{-}hao Su, David Vandyke,
  and Steve~J. Young. 2015.
\newblock Semantically conditioned {LSTM}-based natural language generation for
  spoken dialogue systems.
\newblock In \emph{Empirical Methods in Natural Language Processing (EMNLP)}.

\bibitem[{Weston(2016)}]{DBLL}
Jason Weston. 2016.
\newblock Dialog-based language learning.
\newblock In \emph{Advances in Neural Information Processing Systems (NIPS)}.

\bibitem[{Woodward and Finn(2016)}]{active_oneshot}
Mark Woodward and Chelsea Finn. 2016.
\newblock Active one-shot learning.
\newblock In \emph{NIPS Deep Reinforcement Learning Workshop}.

\bibitem[{Yu et~al.(2018)Yu, Zhang, and Xu}]{Nav}
Haonan Yu, Haichao Zhang, and Wei Xu. 2018.
\newblock Interactive grounded language acquisition and generalization in a
  {2D} world.
\newblock In \emph{International Conference on Learning Representations
  (ICLR)}.

\bibitem[{Yu et~al.(2017)Yu, Zhang, Wang, and Yu}]{SeqGAN}
Lantao Yu, Weinan Zhang, Jun Wang, and Yong Yu. 2017.
\newblock {SeqGAN}: Sequence generative adversarial nets with policy gradient.
\newblock In \emph{Association for the Advancement of Artificial Intelligence
  (AAAI)}.

\bibitem[{Zhang et~al.(2017)Zhang, Yu, and Xu}]{L2T}
Haichao Zhang, Haonan Yu, and Wei Xu. 2017.
\newblock Listen, interact and talk: Learning to speak via interaction.
\newblock In \emph{NIPS Workshop on Visually-Grounded Interaction and
  Language}.

\end{thebibliography}
	\bibliographystyle{acl_natbib}

	\clearpage
	\appendix
	\section{Appendix}
	\label{sec:supplemental}
	
	\begin{figure*}[h]
		\centering
		\begin{overpic}[width = 15cm]{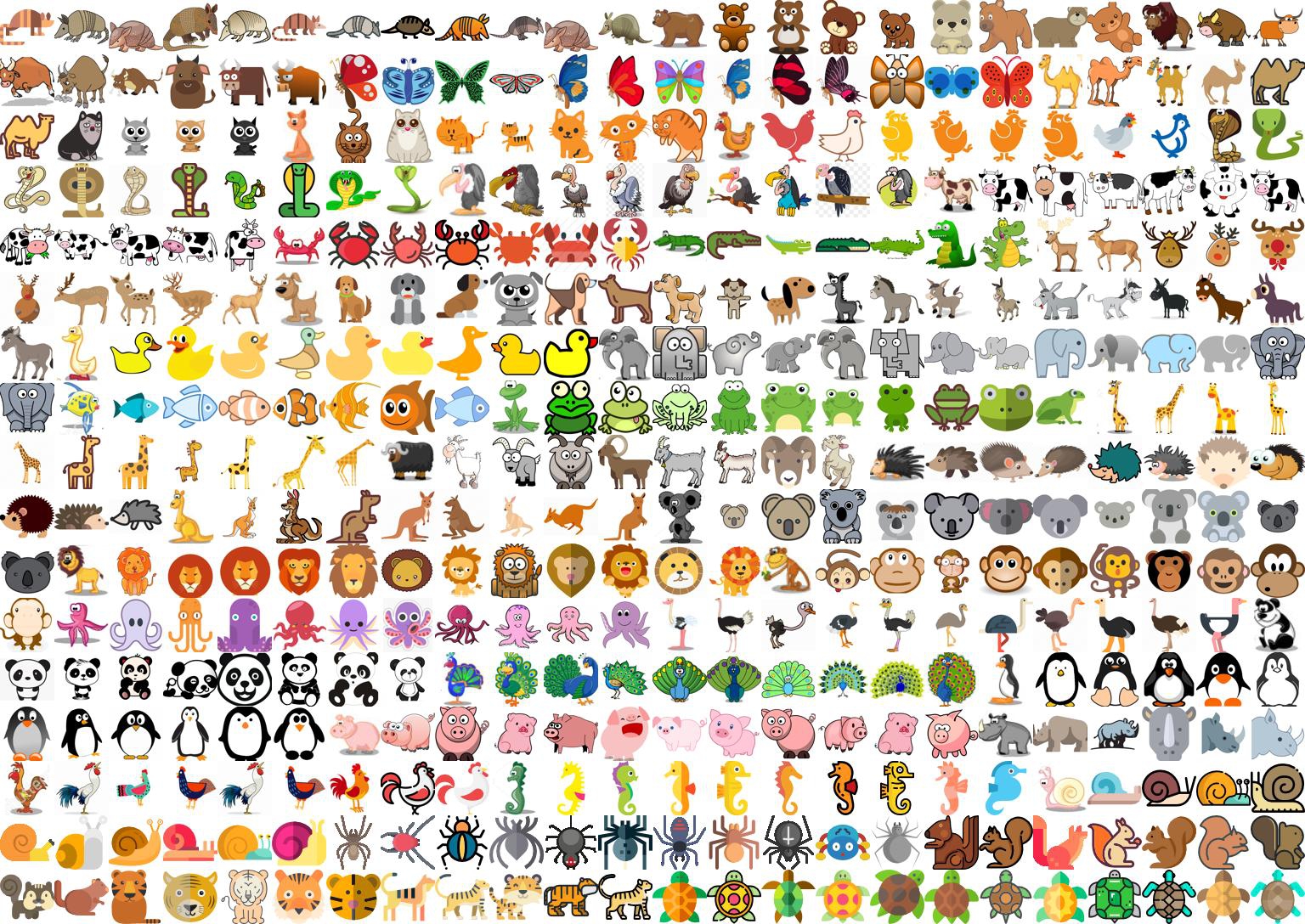}
		\end{overpic}
		\vspace{0.1in}\\
		\begin{overpic}[width = 15cm]{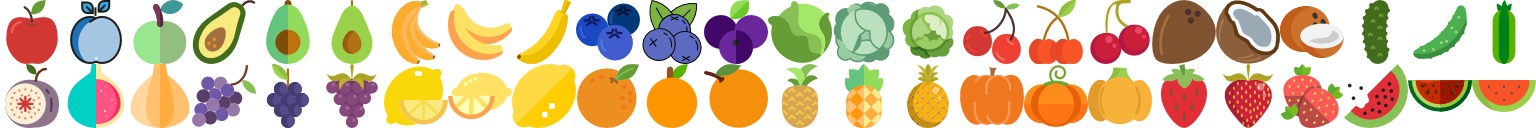}
		\end{overpic}
		\caption{\textbf{Dataset images.} \textbf{Top}: {\footnotesize\textsf{Animal}} dataset. \textbf{Bottom}: {\footnotesize\textsf{Fruit}} dataset.}
		\label{fig:image_mosaic}
	\end{figure*}
	
	\subsection{Datasets and Example Sentences}
	\label{sec:dataset}
	The {\small\textsf{Animal}} dataset contains 40 animal classes with 408 images in total, with about 10 images per class on average.
	The {\small\textsf{Fruit}} dataset contains 16 classes and 48 images in total with 3 images per class. The object classes  and images  are summarized in Table~\ref{tab:obj_class} and Figure~\ref{fig:image_mosaic}. 
	Example sentences from the teacher in different cases (questioning, answering, and saying nothing) are presented in \mbox{Table~\ref{tab:sentence_examples}}.

	\begin{table}[h]
		\footnotesize
		\caption{Object classes for two datasets.} 
		\label{tab:obj_class}
		\vspace{-0.1in} 
		\begin{tabular}[t]{p{0.6cm} |p{1.cm}| p{5.cm} }
			\hline
			Set & \multirow{1}{*}{\footnotesize{\#cls/img}} & \footnotesize{Object Names} \\
			\hline	 
			\multirow{8}{*}{{\scriptsize\textsf{Animal}}}&  \multirow{8}{*}{40/408} &
			\footnotesize{
				armadillo, 
				bear, 
				bull, 
				butterfly, 
				camel, 
				cat, 
				chicken, 
				cobra, 
				condor, 
				cow, 
				crab, 
				crocodile, 
				deer, 
				dog, 
				donkey, 
				duck, 
				elephant, 
				fish, 
				frog, 
				giraffe, 
				goat, 
				hedgehog, 
				kangaroo, 
				koala, 
				lion, 
				monkey, 
				octopus, 
				ostrich, 
				panda, 
				peacock, 
				penguin, 
				pig, 
				rhinoceros, 
				rooster, 
				seahorse, 
				snail, 
				spider, 
				squirrel, 
				tiger, 
				turtle
			} \\
			\hline
			\multirow{4}{*}{{\scriptsize\textsf{Fruit}}}&  \multirow{4}{*}{16/48} & \footnotesize{
				apple, 
				avocado, 
				banana, 
				blueberry, 
				cabbage, 
				cherry, 
				coconut, 
				cucumber, 
				fig, 
				grape, 
				lemon, 
				orange, 
				pineapple, 
				pumpkin, 
				strawberry, 
				watermelon
			} \\
			\hline
		\end{tabular}
	\end{table}

	\begin{table}[h]
		\footnotesize
		\centering
		\caption{{Example sentences from the teacher.}} 
		\label{tab:sentence_examples}
		\vspace{-0.1in} 
		\begin{tabular}[t]{p{1.2cm} | p{5cm} }
			\hline
			Category & Example Sentences\\
			\hline	 \hline
			Empty&  \emph{``''} \\
			\hline
			\multirow{14}{*}{Question} 
			&\emph{``what''} \\
			&\emph{``what is it''} \\
			&\emph{``what is this''}\\
			&\emph{``what is there''}\\
			&\emph{``what do you see''}\\
			&\emph{``what can you see''}\\
			&\emph{``what do you observe''}\\
			&\emph{``what can you observe''}\\
			&\emph{``tell what it is''} \\
			&\emph{``tell what this is''}\\
			&\emph{``tell what there is''}\\
			&\emph{``tell what you see''}\\
			&\emph{``tell what you can see''}\\
			&\emph{``tell what you observe''}\\
			&\emph{``tell what you can observe''}\\
			\hline
			& \emph{``apple''}\\
			\multirow{3}{*}{Answer /}&\emph{``it is apple''}\\
			\multirow{3}{*}{Statement}&\emph{``this is apple''}\\
			&\emph{``there is apple''}\\
			&\emph{``i see apple''}\\
			&\emph{``i observe apple''}\\
			&\emph{``i can see apple''}\\
			&\emph{``i can observe apple''}\\
			\hline
		\end{tabular}
	\end{table}

	\subsection{Network Details}
	\label{sec:network_structure}
	\subsubsection{Visual Encoder}
	The visual encoder takes an input  image and outputs a visual feature vector.
	It is implemented as a convolutional neural network (CNN) followed by fully connected (FC) layers.
	The CNN has four layers. Each layer has 32, 64, 128, 256 filters of size $3\!\times\!3$, followed by max-poolings with a pooling size of 3 and a stride of 2.
	The ReLU activation is used for all layers. 
	Two FC layers with output dimensions of 512 and 1024 are used after the CNN, with ReLU and a linear activations respectively.

	\subsubsection{Interpreter and Speaker}
	\emph{Interpreter} and \emph{speaker} are implemented with interpreter-RNN and speaker-RNN respectively and they share parameters. 
	The RNN is implemented using the Gated Recurrent Unit~\cite{GRU} with a state dimension of $1024$.
	Before inputing to the RNN, word ids are first projected to a word embedding vector of dimension $1024$ followed with two FC layers with ReLU activations and a third FC layer with linear activation, all having output dimensions of $1024$.
	
	\subsubsection{Fusion Gate}
	The fusion gate $g$ is implemented as two FC layers with ReLU activations a third FC layer with a sigmoid activation.
	The output dimensions are $50$, $10$ and $1$ for each layer respectively.

	\subsubsection{Controller}
	The controller $f(\cdot)$ together with the identity mapping forms a residue-structured network as 
	\begin{shrinkeq}{-1.5ex}
		\begin{eqnarray}
		\label{fc_res}
		\begin{split}
		\mathbf{c} =  \mathbf{h} + f(\mathbf{h}).
		\end{split}
		\end{eqnarray}
	\end{shrinkeq}
	$f(\cdot)$  is implemented as two FC layers with ReLU activations and a third FC layer with a linear activation, all having an output dimensions of 1024.
	
	\subsubsection{Value Network}
	The value network is introduced to estimate the expected accumulated future reward.
	It takes the state vector of interpreter-RNN $\mathbf{h}_{\rm I}$ and the confidence $c$ as input.
	It is implemented as two FC layers with ReLU activations and output dimensions of $512$ and $204$ respectively. The third layer is another FC layer with a linear activation and an output dimension of $1$. 
	It is trained by minimizing a cost as~\cite{Sutton}
	\begin{shrinkeq}{-1.ex}
		\begin{eqnarray}
		\nonumber
		\begin{split}
		\mathcal{L}^{\rm V} \!\!=\!\! \mathbb{E}_{p^{\rm S}_{\theta}} \big(V(\mathbf{h}^{t}_{\rm I}, c^t) - r^{t+1}  \!\!-\!\! \lambda V^{'}(\mathbf{h}^{t+1}_{\rm I}, c^{t+1})\big)^2.
		\end{split}
		\end{eqnarray}
	\end{shrinkeq}
	$V^{'}(\cdot)$ denotes a target version of the value network, whose parameters remain fixed until copied from $V(\cdot)$ periodically~\cite{DQN}.
	
	\subsubsection{Confidence Score}
	The confidence score $c$ is defined as follows:
	\begin{shrinkeq}{-1.5ex}
		\begin{eqnarray}
		c\!=\!\max (\mathbf{E}^{\mathsf{T}} \mathbf{r}),
		\end{eqnarray}
	\end{shrinkeq}
	where $\mathbf{E} \!\!\in\!\! \mathbb{R}^{d \times k}$ is the  word embedding table, with  $d$ the embedding dimension  and $k$ the  vocabulary size.
	$\mathbf{r}\!\!\in\!\! \mathbb{R}^{d}$ is the vector read out from the sentence modality of the external memory as:
	\begin{shrinkeq}{-1ex}
		\begin{equation}
		\label{eq:read_alpha}
		\mathbf{r} = \mathbf{M}_s{\boldsymbol\alpha},
		\end{equation}
	\end{shrinkeq}
	where $\boldsymbol{\alpha}$ a soft reading weight  obtained through the visual modality by calculating the cosine similarities between $\mathbf{k}_{v}$ and the slots of $\mathbf{M}_v$.
	The content stored in the memory is extracted from teacher's sentence $\{w_1, w_2, \cdots, w_i, \cdots, w_n\}$ as (detailed in Section~\ref{app:content_selection}):
	\begin{shrinkeq}{-1.ex}
		\begin{eqnarray}
		\label{eq:content}
		\mathbf{c}_{s} = [\boldsymbol{w}_1, \boldsymbol{w}_2, \cdots, \boldsymbol{w}_i \cdots, \boldsymbol{w}_n] \boldsymbol{\eta},
		\end{eqnarray}
	\end{shrinkeq}
	where $\boldsymbol{w}_i\!\!\in\!\! \mathbb{R}^{d}$ denotes the embedding vector extracted from the word embedding table $\mathbf{E}$ for the word $w_i$.
	Therefore, for  a well-learned concept  with effective $\boldsymbol{\eta}$ for information extraction and effective $\boldsymbol{\alpha}$ for information retrieval, $\mathbf{r}$ should be an embedding vector mainly corresponding to the label word associated with the visual image. Therefore, the value of $c$  should be large and the maximum is reached at the location where that label word resides in the embedding table. For a completely novel concept, as the memory contains no information about it, the reading attention $\alpha$ will not be focused 
	and thus $\mathbf{r}$ would be an averaging of a set of existing word embedding vectors in the external memory, leading to a small $c$ value.

	\subsection{Sentence Content Extraction and Importance Gate}
	\label{app:content_selection}
	\subsubsection{Content Extraction}
	We use an attention scheme to extract useful information from a sentence to be written into memory.
	Given a sentence  $\mathbf{w}=\{w_1, w_2, \cdots, w_n\}$
	and the corresponding word embedding vectors $\{\boldsymbol{w}_1, \boldsymbol{w}_2, \cdots, \boldsymbol{w}_n\}$,
	a summary of the sentence is firstly generated using a bidirectional RNN, yielding the states $\{\overrightarrow{\boldsymbol{w}_1}, \overrightarrow{\boldsymbol{w}_2}, \cdots, \overrightarrow{\boldsymbol{w}_n}\}$ for the forward pass and $\{\overleftarrow{\boldsymbol{w}_1}, \overleftarrow{\boldsymbol{w}_2}, \cdots, \overleftarrow{\boldsymbol{w}_n}\}$ for the backward pass. The summary vector is the concatenation of the last state of forward pass and the first state of the backward pass:
	\begin{shrinkeq}{-1.ex}
		\begin{eqnarray}
		\label{eq:sentence_summary}
		\boldsymbol{s} = {\rm concat(\overrightarrow{\boldsymbol{w}_n}, \overleftarrow{\boldsymbol{w}_1})}.
		\end{eqnarray}
	\end{shrinkeq}
	The context vector is the concatenation of the word embedding vector and the state vectors of both forward and backward passes:
	\begin{shrinkeq}{-1.ex}
		\begin{eqnarray}
		\label{eq:context}
		\bar{\boldsymbol{w}}_i = {\rm concat(\boldsymbol{w}_i, \overrightarrow{\boldsymbol{w}_i}, \overleftarrow{\boldsymbol{w}_i})}.
		\end{eqnarray}
	\end{shrinkeq}
	The word level attention $\boldsymbol{\eta}=[\eta_1, \eta_2, \cdots, \eta_i, \cdots]$ is computed as the cosine similarity between transformed sentence summary vector $\boldsymbol{s}$ and each context vector  $\bar{\boldsymbol{w}}_i$:
	\begin{shrinkeq}{-1.ex}
		\begin{eqnarray}
		\label{eq:word_attention}
		{\eta}_i = {\rm cos}\big(f_{\rm MLP}^{\theta_1}(\boldsymbol{s}), f_{\rm MLP}^{\theta_2} (\bar{\boldsymbol{w}}_i)\big).
		\end{eqnarray}
	\end{shrinkeq}
	Both MLPs contain two FC layers with  output dimensions of $1024$ and
	a linear and a Tanh activation for each layer respectively.   
	The content $\mathbf{c}_s$ to be written into the memory is computed as:
	\begin{shrinkeq}{-1.ex}
		\begin{eqnarray}
		\label{eq:content}
		\mathbf{c}_s = \mathbf{W} \boldsymbol{\eta} = [\boldsymbol{w}_1, \boldsymbol{w}_2, \cdots, \boldsymbol{w}_n] \boldsymbol{\eta}.
		\end{eqnarray}
	\end{shrinkeq}

	\subsubsection{Importance Gate}
	The content importance gate is computed as $g_{\rm mem} \!\!=\!\! \sigma(f_{\rm MLP}({\mathbf{s}}))$,
	meaning that the importance of the content  to be written into the memory depends on the sentence from the teacher.
	The MLP contains two FC layers with ReLU activation and output dimensions of $50$ and $30$ respectively.  Another FC layer with a linear activation, and an output dimension of $20$ is used. The output layer is an FC layer with an output dimension of $1$ and a sigmoid activation $\sigma$ .

	\subsection{Example Dialogues on Novel Data}
	\label{sec:example_dialogue}
	We train models on the {\small\textsf{Animal}} dataset and perform  the evaluation on the {\small\textsf{Fruit}} dataset.
	Example dialogues of different approaches are shown in Table~\ref{tab:dialg_example}.
	It is observed that \textbf{Reinforce}  arrives at a policy that the learner keeps silent.
	Both \textbf{Imitation} and \textbf{Imitation+Gaussian-RL}  can generate sensible sentences, but cannot speak adaptively according to context. 
	\textbf{Proposed} can speak according to context adaptively, asking information about novel classes, and answering questions after being taught only once by the teacher.
	
	\begin{table*}[t]
		\centering
		\footnotesize
		\caption{\centering Example dialogues from different approaches.} 
		\label{tab:dialg_example}
		\begin{minipage}[c]{1\textwidth}
			\centering
			\textbf{Reinforce} \\
			\vspace{-0.1in} 
			\begin{tabular}[t]{  c  l  c} 
				\hline
				\multirow{2}{*}{{\fbox{\includegraphics[height=0.5cm]{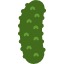}}}} 
				&T: \emph{what} &  \\
				&L:  & \textcolor{red}{\scalebox{.8}{{\cmark}}}	 \\ 
				\hdashline
				\multirow{2}{*}{\fbox{\includegraphics[height=0.5cm]{cucumber_1.jpg}}}
				&T: \emph{i can see cucumber}& \\
				&L: & \textcolor{red}{\scalebox{.8}{{\xmark}}} \\ 		
				\hdashline
				\multirow{2}{*}{\fbox{\includegraphics[height=0.5cm]{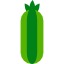}}}
				&T: \emph{there is cucumber} &\\
				&L:  & \textcolor{red}{\scalebox{.8}{{\xmark}}}\\
				\hdashline
				\multirow{2}{*}{\fbox{\includegraphics[height=0.5cm]{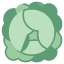}}}
				&T: \emph{tell what  you can observe} &\\
				&L:  & \textcolor{red}{\scalebox{.8}{{\cmark}}}\\
				\hdashline
				\multirow{2}{*}{\fbox{\includegraphics[height=0.5cm]{cabbage_2.jpg}}}
				&T: \emph{i observe cabbage}&\\
				&L:  & \textcolor{red}{\scalebox{.8}{{\xmark}}}\\
				\hdashline
				\multirow{2}{*}{\fbox{\includegraphics[height=0.5cm]{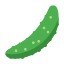}}}
				&T:  &\\
				&L:  & \textcolor{red}{\scalebox{.8}{{\xmark}}}\\		
				\hdashline	 		
				& FAILURE\\
				\hline
			\end{tabular}
			\hspace{0.1in}
			\begin{tabular}[t]{  c  l  c}
				\hline
				\multirow{2}{*}{{\fbox{\includegraphics[height=0.5cm]{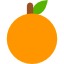}}}} 
				&T: \emph{there is orange} &  \\
				&L:  & \textcolor{red}{\scalebox{.8}{{\xmark}}}	 \\ 
				\hdashline
				\multirow{2}{*}{\fbox{\includegraphics[height=0.5cm]{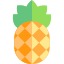}}}
				&T: \emph{tell what it is}& \\
				&L: & \textcolor{red}{\scalebox{.8}{{\cmark}}} \\ 		
				\hdashline
				\multirow{2}{*}{\fbox{\includegraphics[height=0.5cm]{pineapple_2.jpg}}}
				&T: \emph{i see pineapple} &\\
				&L:  & \textcolor{red}{\scalebox{.8}{{\xmark}}}\\
				\hdashline
				\multirow{2}{*}{\fbox{\includegraphics[height=0.5cm]{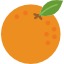}}}
				&T: \emph{what can you see} &\\
				&L:  & \textcolor{red}{\scalebox{.8}{{\cmark}}}\\
				\hdashline
				\multirow{2}{*}{\fbox{\includegraphics[height=0.5cm]{orange_1.jpg}}}
				&T: \emph{there is orange}&\\
				&L:  & \textcolor{red}{\scalebox{.8}{{\xmark}}}\\
				\hdashline
				\multirow{2}{*}{\fbox{\includegraphics[height=0.5cm]{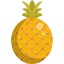}}}
				&T: \emph{what can you see} &\\
				&L:  & \textcolor{red}{\scalebox{.8}{{\xmark}}}\\		
				\hdashline	 		
				& FAILURE\\
				\hline
			\end{tabular}
			\hspace{0.1in}
			\begin{tabular}[t]{  c  l  c}
				\hline
				\multirow{2}{*}{{\fbox{\includegraphics[height=0.5cm]{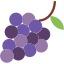}}}} 
				&T: \emph{i observe grape} &  \\
				&L:  & \textcolor{red}{\scalebox{.8}{{\xmark}}}	 \\ 
				\hdashline
				\multirow{2}{*}{\fbox{\includegraphics[height=0.5cm]{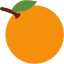}}}
				&T: \emph{i can observe orange}& \\
				&L: & \textcolor{red}{\scalebox{.8}{{\xmark}}} \\ 		
				\hdashline
				\multirow{2}{*}{\fbox{\includegraphics[height=0.5cm]{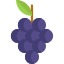}}}
				&T: \emph{what is it} &\\
				&L:  & \textcolor{red}{\scalebox{.8}{{\xmark}}}\\
				\hdashline
				\multirow{2}{*}{\fbox{\includegraphics[height=0.5cm]{grape_2.jpg}}}
				&T: \emph{i see grape} &\\
				&L:  & \textcolor{red}{\scalebox{.8}{{\xmark}}}\\
				\hdashline
				\multirow{2}{*}{\fbox{\includegraphics[height=0.5cm]{orange_1.jpg}}}
				&T: \emph{tell what it is}&\\
				&L:  & \textcolor{red}{\scalebox{.8}{{\xmark}}}\\
				\hdashline
				\multirow{2}{*}{\fbox{\includegraphics[height=0.5cm]{orange_1.jpg}}}
				&T: \emph{i observe orange} &\\
				&L:  & \textcolor{red}{\scalebox{.8}{{\xmark}}}\\		
				\hdashline	 		
				& FAILURE\\
				\hline
			\end{tabular}					
		\end{minipage}
		
		\begin{minipage}[c]{1.1\textwidth} 
			\centering
			\vspace{0.1in}
			\textbf{Imitation} \\
			\vspace{-0.1in} 
			\hspace{-0.6in} 
			\begin{tabular}[t]{  c  l  c}
				\hline
				\multirow{2}{*}{{\fbox{\includegraphics[height=0.5cm]{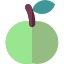}}}} 
				&T: \emph{apple} &  \\
				&L: \emph{what is this} & \textcolor{red}{\scalebox{.8}{{\xmark}}}	 \\ 
				\hdashline
				\multirow{2}{*}{\fbox{\includegraphics[height=0.5cm]{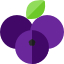}}}
				&T: \emph{tell what it is}& \\
				&L: \emph{i can see condor}& \textcolor{red}{\scalebox{.8}{{\xmark}}} \\ 		
				\hdashline
				\multirow{2}{*}{\fbox{\includegraphics[height=0.5cm]{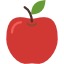}}}
				&T: \emph{this is apple} &\\
				&L: \emph{what is this} & \textcolor{red}{\scalebox{.8}{{\xmark}}}\\
				\hdashline
				\multirow{2}{*}{\fbox{\includegraphics[height=0.5cm]{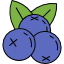}}}
				&T: \emph{tell what there is} &\\
				&L: \emph{i can see apple} & \textcolor{red}{\scalebox{.8}{{\xmark}}}\\
				\hdashline
				\multirow{2}{*}{\fbox{\includegraphics[height=0.5cm]{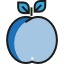}}}
				&T:  &\\
				&L: \emph{what is this} & \textcolor{red}{\scalebox{.8}{{\xmark}}}\\
				\hdashline
				\multirow{2}{*}{\fbox{\includegraphics[height=0.5cm]{apple_2.jpg}}}
				&T: \emph{i can observe apple} &\\
				&L: \emph{what is this} & \textcolor{red}{\scalebox{.8}{{\xmark}}}\\		
				\hdashline	 		
				& FAILURE\\
				\hline
			\end{tabular}
			\hspace{0.08in}
			\begin{tabular}[t]{  c  l  c}
				\hline
				\multirow{2}{*}{{\fbox{\includegraphics[height=0.5cm]{apple_1.jpg}}}} 
				&T: \emph{what is there} &  \\
				&L: \emph{i can observe condor} & \textcolor{red}{\scalebox{.8}{{\xmark}}}	 \\ 
				\hdashline
				\multirow{2}{*}{\fbox{\includegraphics[height=0.5cm]{cucumber_2.jpg}}}
				&T: \emph{tell what you see}& \\
				&L: \emph{i can see condor}& \textcolor{red}{\scalebox{.8}{{\xmark}}} \\ 		
				\hdashline
				\multirow{2}{*}{\fbox{\includegraphics[height=0.5cm]{apple_2.jpg}}}
				&T: \emph{tell what you can see} &\\
				&L: \emph{i can see condor} & \textcolor{red}{\scalebox{.8}{{\xmark}}}\\
				\hdashline
				\multirow{2}{*}{\fbox{\includegraphics[height=0.5cm]{apple_3.jpg}}}
				&T: \emph{what is there} &\\
				&L: \emph{i can see condor} & \textcolor{red}{\scalebox{.8}{{\xmark}}}\\
				\hdashline
				\multirow{2}{*}{\fbox{\includegraphics[height=0.5cm]{cucumber_1.jpg}}}
				&T:  &\\
				&L: \emph{what is this} & \textcolor{red}{\scalebox{.8}{{\cmark}}}\\
				\hdashline
				\multirow{2}{*}{\fbox{\includegraphics[height=0.5cm]{cucumber_1.jpg}}}
				&T: \emph{i can see cucumber} &\\
				&L: \emph{what is this} & \textcolor{red}{\scalebox{.8}{{\xmark}}}\\		
				\hdashline	 		
				& FAILURE\\
				\hline
			\end{tabular}
			\hspace{0.08in}	
			\begin{tabular}[t]{  c  l  c}
				\hline
				\multirow{2}{*}{{\fbox{\includegraphics[height=0.5cm]{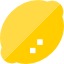}}}} 
				&T: \emph{i see lemon} &  \\
				&L: \emph{what is this} & \textcolor{red}{\scalebox{.8}{{\xmark}}}	 \\ 
				\hdashline
				\multirow{2}{*}{\fbox{\includegraphics[height=0.5cm]{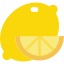}}}
				&T: \emph{i see lemon}& \\
				&L: \emph{what is this}& \textcolor{red}{\scalebox{.8}{{\xmark}}} \\ 		
				\hdashline
				\multirow{2}{*}{\fbox{\includegraphics[height=0.5cm]{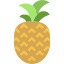}}}
				&T: \emph{tell what you can observe} &\\
				&L: \emph{i can see lemon} & \textcolor{red}{\scalebox{.8}{{\xmark}}}\\
				\hdashline
				\multirow{2}{*}{\fbox{\includegraphics[height=0.5cm]{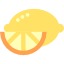}}}
				&T: \emph{tell what there is} &\\
				&L: \emph{what is this} & \textcolor{red}{\scalebox{.8}{{\xmark}}}\\
				\hdashline
				\multirow{2}{*}{\fbox{\includegraphics[height=0.5cm]{lemon_2.jpg}}}
				&T: \emph{it is lemon}  &\\
				&L: \emph{what is this} & \textcolor{red}{\scalebox{.8}{{\xmark}}}\\
				\hdashline
				\multirow{2}{*}{\fbox{\includegraphics[height=0.5cm]{pineapple_3.jpg}}}
				&T: \emph{tell what you observe} &\\
				&L: \emph{what is this} & \textcolor{red}{\scalebox{.8}{{\xmark}}}\\		
				\hdashline	 		
				& FAILURE\\
				\hline
			\end{tabular}
		\end{minipage}

		\begin{minipage}[c]{1.1\textwidth} 
			\vspace{0.1in}
			\centering
			\textbf{Imitation+Gaussian-RL} \\
			\vspace{-0.1in}
			\hspace{-0.5in} 
			\begin{tabular}[t]{  c  l  c}
				\hline
				\multirow{2}{*}{{\fbox{\includegraphics[height=0.5cm]{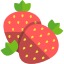}}}} 
				&T: \emph{i can see strawberry} &  \\
				&L:  \emph{i can see snail} & \textcolor{red}{\scalebox{.8}{{\xmark}}}	 \\ 
				\hdashline
				\multirow{2}{*}{\fbox{\includegraphics[height=0.5cm]{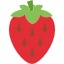}}}
				&T: \emph{it is strawberry}& \\
				&L: \emph{i can see strawberry} & \textcolor{red}{\scalebox{.8}{{\cmark}}} \\ 		
				\hdashline
				\multirow{2}{*}{\fbox{\includegraphics[height=0.5cm]{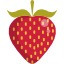}}}
				&T: \emph{i can see strawberry} &\\
				&L:  \emph{i can see strawberry} & \textcolor{red}{\scalebox{.8}{{\cmark}}}\\
				\hdashline
				\multirow{2}{*}{\fbox{\includegraphics[height=0.5cm]{blueberry_2.jpg}}}
				&T: \emph{blueberry} &\\
				&L:  \emph{i can see blueberry} & \textcolor{red}{\scalebox{.8}{{\cmark}}}\\	
				\hdashline	 
				\multirow{2}{*}{\fbox{\includegraphics[height=0.5cm]{blueberry_3.jpg}}}
				&T: \emph{what} &\\
				&L: \emph{i can see blueberry} & \textcolor{red}{\scalebox{.8}{{\cmark}}}\\	
				\hdashline		
				& FAILURE\\
				\hline
			\end{tabular}
			\hspace{0.01in}		
			\begin{tabular}[t]{  c  l  c}
				\hline
				\multirow{2}{*}{{\fbox{\includegraphics[height=0.5cm]{lemon_3.jpg}}}} 
				&T:  \emph{what do you observe}  &  \\
				&L: \emph{i can see tiger} & \textcolor{red}{\scalebox{.8}{{\xmark}}}	 \\ 
				\hdashline
				\multirow{2}{*}{\fbox{\includegraphics[height=0.5cm]{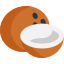}}}
				&T: \emph{i can observe coconut}& \\
				&L: \emph{i can see coconut} & \textcolor{red}{\scalebox{.8}{{\cmark}}} \\ 		
				\hdashline
				\multirow{2}{*}{\fbox{\includegraphics[height=0.5cm]{lemon_1.jpg}}}
				&T: \emph{what can you see} &\\
				&L:  \emph{i can see tiger} & \textcolor{red}{\scalebox{.8}{{\xmark}}}\\
				\hdashline
				\multirow{2}{*}{\fbox{\includegraphics[height=0.5cm]{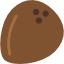}}}
				&T: \emph{what} &\\
				&L: \emph{i can see dog} & \textcolor{red}{\scalebox{.8}{{\xmark}}}\\	
				\hdashline	 
				\multirow{2}{*}{\fbox{\includegraphics[height=0.5cm]{lemon_2.jpg}}}
				&T: \emph{tell what  you can observe} &\\
				&L: \emph{i can see tiger} & \textcolor{red}{\scalebox{.8}{{\xmark}}}\\	
				\hdashline		
				& FAILURE\\
				\hline
			\end{tabular}
			\hspace{0.01in}
			\begin{tabular}[t]{  c  l  c}
				\hline
				\multirow{2}{*}{{\fbox{\includegraphics[height=0.5cm]{strawberry_2.jpg}}}} 
				&T:  \emph{i can observe strawberry} &  \\
				&L: \emph{i can see strawberry} & \textcolor{red}{\scalebox{.8}{{\cmark}}}	 \\ 
				\hdashline
				\multirow{2}{*}{\fbox{\includegraphics[height=0.5cm]{pineapple_1.jpg}}}
				&T: \emph{this is pineapple}& \\
				&L: \emph{i can see pineapple}& \textcolor{red}{\scalebox{.8}{{\cmark}}} \\ 		
				\hdashline
				\multirow{2}{*}{\fbox{\includegraphics[height=0.5cm]{pineapple_2.jpg}}}
				&T: \emph{this is pineapple} &\\
				&L: \emph{i can see pineapple} & \textcolor{red}{\scalebox{.8}{{\cmark}}}\\
				\hdashline
				\multirow{2}{*}{\fbox{\includegraphics[height=0.5cm]{strawberry_3.jpg}}}
				&T: \emph{what do you see} &\\
				&L: \emph{i can see pineapple} & \textcolor{red}{\scalebox{.8}{{\xmark}}}\\	
				\hdashline	 		
				& FAILURE\\
				\hline
			\end{tabular}
		\end{minipage}\\
		\begin{minipage}[c]{1.01\textwidth} 
			\vspace{0.1in}
			\centering
			\textbf{Proposed} \\
			\vspace{-0.1in} 
			\hspace{-0.1in} 
			\begin{tabular}[t]{  c  l  c}
				\hline
				\multirow{2}{*}{{\fbox{\includegraphics[height=0.5cm]{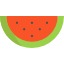}}}} 
				&T: \emph{} &  \\
				&L: \emph{tell what you observe} & \textcolor{red}{\scalebox{.8}{{\cmark}}}	 \\ 
				\hdashline
				\multirow{2}{*}{\fbox{\includegraphics[height=0.5cm]{watermelon_3.jpg}}}
				&T: \emph{this is watermelon}& \\
				&L: \emph{watermelon}& \textcolor{red}{\scalebox{.8}{{\cmark}}} \\ 		
				\hdashline
				\multirow{2}{*}{\fbox{\includegraphics[height=0.5cm]{lemon_3.jpg}}}
				&T: \emph{tell what there is} &\\
				&L: \emph{tell what you see} & \textcolor{red}{\scalebox{.8}{{\cmark}}}\\			
				\hdashline
				\multirow{2}{*}{\fbox{\includegraphics[height=0.5cm]{lemon_3.jpg}}}
				&T: \emph{i observe lemon} &\\
				&L: \emph{lemon} & \textcolor{red}{\scalebox{.8}{{\cmark}}}\\	
				\hdashline
				\multirow{2}{*}{\fbox{\includegraphics[height=0.5cm]{lemon_1.jpg}}}
				&T: \emph{} &\\
				&L: \emph{lemon} & \textcolor{red}{\scalebox{.8}{{\cmark}}}\\
				\hdashline		
				& SUCCESS\\
				\hline
			\end{tabular}
			\hspace{0.08in}
			\begin{tabular}[t]{  c  l  c}
				\hline
				\multirow{2}{*}{{\fbox{\includegraphics[height=0.5cm]{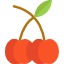}}}} 
				&T: \emph{tell what you can observe} &  \\
				&L: \emph{tell what you observe} & \textcolor{red}{\scalebox{.8}{{\cmark}}}	 \\ 
				\hdashline
				\multirow{2}{*}{\fbox{\includegraphics[height=0.5cm]{cherry_2.jpg}}}
				&T: \emph{i observe cherry}& \\
				&L: \emph{cherry}& \textcolor{red}{\scalebox{.8}{{\cmark}}} \\ 		
				\hdashline
				\multirow{2}{*}{\fbox{\includegraphics[height=0.5cm]{orange_3.jpg}}}
				&T: \emph{i observe orange}&\\
				&L: \emph{orange} & \textcolor{red}{\scalebox{.8}{{\cmark}}}\\			
				\hdashline	
				\multirow{2}{*}{\fbox{\includegraphics[height=0.5cm]{orange_1.jpg}}}
				&T: \emph{tell what there is}& \\
				&L: \emph{orange}& \textcolor{red}{\scalebox{.8}{{\cmark}}} \\ 					
				\hdashline		
				& SUCCESS\\
				\hline
			\end{tabular}
			\hspace{0.08in}
			\begin{tabular}[t]{  c  l  c}
				\hline
				\multirow{2}{*}{{\fbox{\includegraphics[height=0.5cm]{blueberry_3.jpg}}}} 
				&T: \emph{} &  \\
				&L: \emph{tell what you observe} & \textcolor{red}{\scalebox{.8}{{\cmark}}}	 \\ 
				\hdashline
				\multirow{2}{*}{\fbox{\includegraphics[height=0.5cm]{blueberry_3.jpg}}}
				&T: \emph{i see blueberry}& \\
				&L: \emph{blueberry}& \textcolor{red}{\scalebox{.8}{{\cmark}}} \\ 		
				\hdashline
				\multirow{2}{*}{\fbox{\includegraphics[height=0.5cm]{orange_2.jpg}}}
				&T: \emph{there is orange} &\\
				&L: \emph{orange} & \textcolor{red}{\scalebox{.8}{{\cmark}}}\\			
				\hdashline
				\multirow{2}{*}{\fbox{\includegraphics[height=0.5cm]{orange_3.jpg}}}
				&T: \emph{i can observe orange} &\\
				&L: \emph{orange} & \textcolor{red}{\scalebox{.8}{{\cmark}}}\\	
				\hdashline
				\multirow{2}{*}{\fbox{\includegraphics[height=0.5cm]{blueberry_2.jpg}}}
				&T: \emph{tell what it is} &\\
				&L: \emph{this is blueberry} & \textcolor{red}{\scalebox{.8}{{\cmark}}}\\
				\hdashline	
				& SUCCESS\\
				\hline
			\end{tabular} \\
			\hspace{-0.1in}
			\begin{tabular}[t]{  c  l  c}
				\hline
				\multirow{2}{*}{{\fbox{\includegraphics[height=0.5cm]{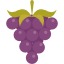}}}} 
				&T: \emph{i can see grape} &  \\
				&L: \emph{grape} & \textcolor{red}{\scalebox{.8}{{\cmark}}}	 \\ 
				\hdashline
				\multirow{2}{*}{\fbox{\includegraphics[height=0.5cm]{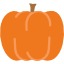}}}
				&T: \emph{}& \\
				&L: \emph{tell what you observe}& \textcolor{red}{\scalebox{.8}{{\cmark}}} \\ 		
				\hdashline
				\multirow{2}{*}{\fbox{\includegraphics[height=0.5cm]{pumpkin_1.jpg}}}
				&T: \emph{i can see pumpkin}& \\
				&L: \emph{pumpkin}& \textcolor{red}{\scalebox{.8}{{\cmark}}} \\ 		
				\hdashline
				\multirow{2}{*}{\fbox{\includegraphics[height=0.5cm]{grape_2.jpg}}}
				&T: \emph{} &\\
				&L: \emph{grape} & \textcolor{red}{\scalebox{.8}{{\cmark}}}\\
				\hdashline		 		
				& SUCCESS\\
				\hline
			\end{tabular}
			\hspace{0.08in}
			\begin{tabular}[t]{  c  l  c}
				\hline
				\multirow{2}{*}{{\fbox{\includegraphics[height=0.5cm]{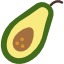}}}} 
				&T: \emph{what} &  \\
				&L: \emph{tell what you observe} & \textcolor{red}{\scalebox{.8}{{\cmark}}}	 \\ 
				\hdashline
				\multirow{2}{*}{\fbox{\includegraphics[height=0.5cm]{avocado_1.jpg}}}
				&T: \emph{i can see avocado}& \\
				&L: \emph{avocado}& \textcolor{red}{\scalebox{.8}{{\cmark}}} \\ 		
				\hdashline
				\multirow{2}{*}{\fbox{\includegraphics[height=0.5cm]{coconut_3.jpg}}}
				&T: \emph{this is coconut} &\\
				&L: \emph{coconut} & \textcolor{red}{\scalebox{.8}{{\cmark}}}\\			
				\hdashline
				\multirow{2}{*}{\fbox{\includegraphics[height=0.5cm]{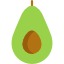}}}
				&T: \emph{what is there} &\\
				&L: \emph{this is avocado} & \textcolor{red}{\scalebox{.8}{{\cmark}}}\\	
				\hdashline	
				& SUCCESS\\
				\hline
			\end{tabular}
			\hspace{0.08in}
			\begin{tabular}[t]{  c  l  c}
				\hline
				\multirow{2}{*}{{\fbox{\includegraphics[height=0.5cm]{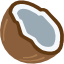}}}} 
				&T: \emph{it is coconut}&  \\
				&L: \emph{i can see coconut} & \textcolor{red}{\scalebox{.8}{{\cmark}}}	 \\ 
				\hdashline
				\multirow{2}{*}{\fbox{\includegraphics[height=0.5cm]{pumpkin_1.jpg}}}
				&T: \emph{this is pumpkin}& \\
				&L: \emph{i can see pumpkin}& \textcolor{red}{\scalebox{.8}{{\cmark}}} \\ 		
				\hdashline
				\multirow{2}{*}{\fbox{\includegraphics[height=0.5cm]{coconut_1.jpg}}}
				&T:  \emph{what is it} &\\
				&L: \emph{i can see pumpkin} & \textcolor{red}{\scalebox{.8}{{\xmark}}}\\			
				\hdashline
				\multirow{2}{*}{\fbox{\includegraphics[height=0.5cm]{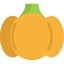}}}
				&T: \emph{what do you see} &\\
				&L: \emph{i can see pumpkin} & \textcolor{red}{\scalebox{.8}{{\cmark}}}\\
				\hdashline		
				& FAILURE\\
				\hline
			\end{tabular}
			
		\end{minipage}		
		
	\end{table*}
	
\end{document}